\documentclass[11pt]{article}
\usepackage[preprint]{acl}

\usepackage{times}
\usepackage{latexsym}
\usepackage[T1]{fontenc}
\usepackage[utf8]{inputenc}
\usepackage{microtype}
\usepackage{graphicx}
\usepackage{fontawesome5}

\usepackage{xspace}
\newif\ifshowcomments
\showcommentstrue
\makeatletter
\let\todo\@undefined
\makeatother

\ifshowcomments
    \usepackage[textsize=scriptsize,textwidth=2.4cm]{todonotes}
\else
    \usepackage[disable,textsize=scriptsize,textwidth=2.4cm]{todonotes}
\fi

\usepackage{amsmath,amsfonts,bm}

\def\eqref#1{equation~\ref{#1}}
\def\1{\bm{1}}

\DeclareMathAlphabet{\mathsfit}{\encodingdefault}{\sfdefault}{m}{sl}
\SetMathAlphabet{\mathsfit}{bold}{\encodingdefault}{\sfdefault}{bx}{n}

\usepackage{bbm}
\usepackage{tikz}
\usetikzlibrary{arrows.meta, positioning, fit, backgrounds, calc}
\usepackage{soul}
\usepackage{ulem}
\usepackage{lipsum}
\usepackage{amssymb}
\usepackage{amsmath}
\usepackage{xspace}
\usepackage{float}
\usepackage{multirow}
\usepackage{booktabs}
\newcommand{\ci}[2]{{\,\tiny[#1,\,#2]}}   % 95% CI in table cells
\usepackage{colortbl}
\usepackage{enumitem}
\usepackage{algpseudocode}
\usepackage{subcaption}
\usepackage{xcolor}
\usepackage{caption}
\usepackage{array}
\usepackage{makecell}
\usepackage[noupquote]{inconsolata}
\usepackage[ruled,vlined]{algorithm2e}
\usepackage[framemethod=TikZ]{mdframed}
\newcommand{\SmallHeading}[1]{\noindent\textbf{#1}.}

\newcommand{\Figure}{Fig.\xspace}
\newcommand{\Figures}{Figs.\xspace}
\newcommand{\Table}{Tab.\xspace}
\newcommand{\Section}{Sec.\xspace}
\newcommand{\Appendix}{App.\xspace}

\definecolor{mypositive}{RGB}{0, 128, 0}
\definecolor{mynegative}{RGB}{220, 20, 60}
\definecolor{mypositive}{RGB}{24, 103, 173}
\definecolor{mynegative}{RGB}{255, 128, 0}

\newcommand{\grayColor}[0]{gray!10}

\definecolor{customblue}{HTML}{1F77B4}
\definecolor{customorange}{HTML}{FF7F0E}
\definecolor{customgreen}{HTML}{2CA02C}
\definecolor{customred}{HTML}{D62728}
\definecolor{maincyan}{RGB}{156,212,228}
\definecolor{navy}{RGB}{0,65,130}
\definecolor{inkdark}{RGB}{26,42,58}
\definecolor{actblue}{RGB}{0,62,134}
\definecolor{actbg}{RGB}{215,235,253}
\definecolor{obsred}{RGB}{165,52,26}
\definecolor{obsbg}{RGB}{255,237,215}
\definecolor{tracebg}{RGB}{240,245,249}
\definecolor{fmtorange}{RGB}{238,129,46}
\definecolor{fmtblue}{RGB}{19,113,172}
\definecolor{membg}{RGB}{225,232,240}
\definecolor{taskink}{HTML}{1E3A8A}
\definecolor{colhead}{HTML}{334155}
\definecolor{sharedbg}{HTML}{FFEDD5}
\definecolor{sharedink}{HTML}{9A3412}
\definecolor{plainbg}{HTML}{F8FAFC}
\definecolor{plainink}{HTML}{64748B}
\newcommand{\actpill}[1]{\tikz[baseline=(P.base)]\node[rounded corners=2pt, inner sep=2pt, minimum width=0pt, minimum height=0pt, fill=actbg, text=actblue, font=\footnotesize] (P) {#1};}
\newcommand{\obspill}[1]{\tikz[baseline=(P.base)]\node[rounded corners=2pt, inner sep=2pt, minimum width=0pt, minimum height=0pt, fill=obsbg, text=obsred, font=\footnotesize] (P) {#1};}
\newcommand{\figicon}[2][3.2mm]{\raisebox{-0.2\height}{\includegraphics[height=#1]{#2}}}

\newcolumntype{C}[1]{>{\centering\let\newline\\\arraybackslash\hspace{0pt}}m{#1}}

\usepackage[skins,breakable]{tcolorbox}

\newtcolorbox{boxblue}{enhanced,colback=blue!5!white,colframe=blue!75!black,breakable=true}

\newtcolorbox{boxsystem}{
  enhanced,
  colback=blue!5!white,
  colframe=blue!75!black,
  breakable=true,
  fontupper=\ttfamily
}

\newtcolorbox{boxuser}{
  enhanced,
  colback=yellow!5!white,
  colframe=yellow!75!black,
  breakable=true,
  fontupper=\ttfamily,
}

\usepackage{titletoc}
\usepackage{listings}
\lstdefinestyle{promptverb}{
  basicstyle=\ttfamily\scriptsize,
  breaklines=true,
  breakindent=0pt,
  breakautoindent=false,
  postbreak=\mbox{\textcolor{gray}{\tiny$\hookrightarrow$}},
  keepspaces=true,
  columns=flexible,
  aboveskip=0pt,
  belowskip=0pt,
  literate={—}{{---}}1 {→}{{$\rightarrow$}}1,
}
\newcommand{\promptfile}[1]{\lstinputlisting[style=promptverb]{prompts/#1}}
\newtcolorbox{promptsystem}[1][System]{enhanced,colback=blue!5!white,colframe=blue!75!black,boxrule=0.6pt,arc=2mm,left=5pt,right=5pt,top=4pt,bottom=4pt,before upper={{\rmfamily\bfseries\small\color{blue!50!black}#1}\par\vspace{2pt}}}
\newtcolorbox{promptuser}[1][User]{enhanced,colback=yellow!8!white,colframe=yellow!60!black,boxrule=0.6pt,arc=2mm,left=5pt,right=5pt,top=4pt,bottom=4pt,before upper={{\rmfamily\bfseries\small\color{yellow!40!black}#1}\par\vspace{2pt}}}
\newtcolorbox{skillbox}[1][Skill]{enhanced,colback=customgreen!6!white,colframe=customgreen!60!black,boxrule=0.6pt,arc=2mm,left=5pt,right=5pt,top=4pt,bottom=4pt,before upper={{\rmfamily\bfseries\small\color{customgreen!45!black}#1}\par\vspace{2pt}}}

\usepackage{silence}
\title{Break It Down, Pass It On: Cross-Task Skill Transfer in LLM Agents}

\author{%
  \textbf{Yiyang Feng}\quad
  \textbf{Biddut Sarker Bijoy}\quad
  \textbf{Niranjan Balasubramanian}\quad
  \textbf{Jiawei Zhou} \\
  Computer Science Department, Stony Brook University, Stony Brook, USA \\
  \texttt{\{yiyfeng, bbijoy, niranjan, jiawei.zhou.1\}@cs.stonybrook.edu}\\[0.5em]
  {\normalsize
   \href{https://github.com/Zesearch/skill-transfer-llm-agents}{\color{black}\faGithub\ \texttt{GitHub Repository}}%
   \hspace{2em}%
   \href{https://zesearch.github.io/subtask-skill-transfer/}{\color{black}\faGlobe\ \texttt{Website Page}}}%
}

\begin{document}
\maketitle

% Abstract
\begin{abstract}
Large language model (LLM) agents can induce skills from completed tasks and reuse them later to grow more capable with experience. In practice, induced skills may transfer unreliably and can even harm the agent that retrieves them. When agent-induced skills transfer reliably across tasks remains an open question. We conduct a comprehensive and controlled study of how the way skills are induced shapes their transfer across tasks. 
% \jz{that xxx, where xxx is the general goal, instead of the detailed control yet. Otherwise it is not very easy to undestand. E.g. text vs. code skill, etc. are really just in our paper control, not universal concepts that readers would immediate recognize.} 
Specifically,
% \jz{from this sentence on (the part just before), try to break down the sentence to be clearer. This sentence mixes in many things without proper context. Try to say the high-level first, and then introduce the controlled axes.} 
we compare task-level with subtask-level skill induction and text with code skill formats, the two axes along which existing methods differ.
Task-level skills mostly reduce the agent's performance
% \jz{rephrase the expression of this sentence to be slighly formal} 
below its no-memory baseline while subtask-level skills raise it above on average, and text skills transfer better than code skills. 
% \jz{I might say sth like: (to frame this as a clear 2nd contribution that we come up with) To further understand/explain/ground our findings, we propose to examine ... two complimentary ... of ... they show complimentary predictive power of task successes... and propose a \textit{skill utility} score. The score shows consistent correlation with task success when skills are being transfered and used, and its computation does not require task execution, providing a practical diagnoistic tool to assess a skill ... before any task runs.} 
To further understand our findings, we examine two complementary properties of the induced skills: \textit{specificity},
% \jz{this sentence for the two metrics reads a bit rushing; try to place specificity and attractiveness clearly two parts easy to read, e.g. "... specificity, which xxx, and abstractiveness, which xxx."} 
which measures how closely a skill matches real tasks, and \textit{abstractness}, which measures how evenly its relevance spreads across tasks. Neither property alone predicts task success, but their combined effect does, which we propose as a \textit{skill utility} score. 
The score correlates consistently with task success when skills are transferred, and subtask-level and text skills score higher. 
% It will be better to remove the claim "no task execution" and "before new task runs" because actually we need task executed to produce skills, and we need to run on a full test set. So it is a post-hoc approach.
Computing skill utility only needs the skills and task descriptions, so our score serves as a lightweight diagnostic of a skill memory for practitioners.
% {\interfootnotelinepenalty10000\footnote{Code and data are available at 
% \url{https://anonymous.4open.science/r/subtask-skill-transfer-68E8}.}}
% \url{https://github.com/Zesearch/skill-transfer-llm-agents}.}}
% Consistent with our findings, subtask-level and text skills have higher utility. Our metric needs no task execution, so practitioners can check a skill memory before any task runs.

\end{abstract}

\section{Introduction}
\label{sec:intro}

\begin{figure}[t]
\centering
\resizebox{\columnwidth}{!}{%
\fontfamily{ppl}\selectfont
\begin{tikzpicture}[
  font=\small,
  rlab/.style={font=\small\bfseries, text=taskink},
  band/.style={font=\scriptsize\bfseries, text=colhead, align=left},
  bandnote/.style={font=\scriptsize, text=plainink, align=left},
  desc/.style={font=\footnotesize, text=colhead, text width=29mm, align=left, inner sep=0pt, execute at begin node={\hyphenpenalty=10000\relax}},
  cell/.style={rounded corners=3pt, align=center, inner sep=2.5pt, minimum width=31mm},
  tskill/.style={cell, font=\footnotesize, minimum height=8mm},
  spill/.style={cell, fill=sharedbg, draw=sharedink!40, text=sharedink, font=\footnotesize\bfseries, minimum height=5.5mm},
  ppill/.style={cell, font=\footnotesize, minimum height=5.5mm},
  xink/.style={fill=taskink!8, draw=taskink!40, text=taskink},
  yink/.style={fill=customgreen!10, draw=customgreen!55, text=customgreen!55!black},
  rule/.style={draw=black!15, semithick},
]
\def\CX{3.6} \def\CY{6.9}
% ---- column headers ----
\node[rlab] at (\CX,5.75) {Task X};
\node[rlab] at (\CY,5.75) {Task Y};
\draw[rule] (0.1,5.47) -- (8.5,5.47);
% ---- task descriptions ----
\node[desc, anchor=north] at (\CX,5.25) {Place an order for all weightlifting benches in my Amazon cart.};
\node[desc, anchor=north] at (\CY,5.25) {Move all food processors from my Amazon cart to wish list.};
\draw[rule] (0.1,3.8) -- (8.5,3.8);
% ---- task-level band ----
\node[band, anchor=west] at (0.05,3.35) {Task-Level\\ Skills};
\node[bandnote, anchor=west] at (0.05,2.92) {not shared};
\node[tskill, xink] at (\CX,3.25) {Order all \{product\}\\ in Amazon cart};
\node[tskill, yink] at (\CY,3.25) {Move all \{product\}\\ to wish list};
\draw[rule] (0.1,2.65) -- (8.5,2.65);
% ---- subtask-level band ----
\node[band, anchor=west] at (0.05,1.6) {Subtask-Level\\ Skills};
\node[bandnote, anchor=west] at (0.05,1.15) {mostly shared};
\node[spill] at (\CX,2.25) {Login};
\node[spill] at (\CY,2.25) {Login};
\node[spill] at (\CX,1.6) {Check Cart};
\node[spill] at (\CY,1.6) {Check Cart};
\node[ppill, xink] at (\CX,0.95) {Place Order};
\node[ppill, yink] at (\CY,0.95) {Move to Wish List};
\node[spill] at (\CX,0.3) {Remove Cart Items};
\node[spill] at (\CY,0.3) {Remove Cart Items};
\draw[rule] (0.1,-0.15) -- (8.5,-0.15);
% ---- legend ----
\draw[rounded corners=1pt, fill=sharedbg, draw=sharedink!40] (1.0,-0.66) rectangle (1.32,-0.44);
\node[font=\scriptsize, text=colhead, anchor=west] at (1.4,-0.55) {shared skill};
\draw[rounded corners=1pt, fill=taskink!8, draw=taskink!40] (3.1,-0.66) rectangle (3.42,-0.44);
\node[font=\scriptsize, text=colhead, anchor=west] at (3.5,-0.55) {Task X only};
\draw[rounded corners=1pt, fill=customgreen!10, draw=customgreen!55] (5.2,-0.66) rectangle (5.52,-0.44);
\node[font=\scriptsize, text=colhead, anchor=west] at (5.6,-0.55) {Task Y only};
\end{tikzpicture}%
}
\caption{Different tasks share subtasks. A task-level skill summarizes the whole trajectory, so it stays tied to its source task and is not shared with the other task. Subtask-level induction yields one skill per subtask, and the skills of shared subtasks (amber) transfer between Task X and Task Y.}
\label{fig:intro-illustration}
\end{figure}

% Stakes. Agents do not improve with experience, and cross-task skill transfer would change that.
Despite the wide use of large language model (LLM) agents in workflows such as personal assistance, office automation, and scientific research \cite{trivedi-etal-2024-appworld, wang2024officebench, lai2025kramabench}, these agents in their naive form solve each task from scratch and do not get better with experience. A growing line of work enables agents to induce skills from completed tasks, store them in a skill memory, and retrieve them for later tasks \citep{wang2024voyager, Zhao_Huang_Xu_Lin_Liu_Huang_2024, wang2025agent, wang2025inducing}. If the induced skills transfer across tasks, an agent becomes increasingly capable over time.

% Challenge. Full-trajectory skills do not scale, with documented failure modes, so the extraction question is open.
In practice, however, skill transfer is unreliable when skills are induced from the whole trajectories of completed tasks. First, a skill induced from a whole trajectory is tied to its source task (\Figure~\ref{fig:intro-illustration}) and generalizes poorly to new tasks \citep{yu2025polyskill, fang2026mempexploringagentprocedural}. Second, these over-specialized skills enter later tasks as irrelevant or misaligned context, which distracts the model and propagates errors from the source task \citep{pmlr-v202-shi23a, yoran2024making, xiong-etal-2026-memory}. Therefore, whether skill transfer helps depends on how skills are induced, raising our central question: \textit{when do agent-induced skills transfer reliably across tasks?}

% Subtask-level induction has emerged, but the evidence is narrow, so a fair and general analysis is needed.
A recent line of work suggests inducing skills at the subtask level
% \jz{maybe complete the sentence? suggests inducing skills on the subtask level.}
% subtask-level skill induction 
\citep{nottingham2024skill, yang-etal-2026-towards, shen2026structurallyalignedsubtasklevelmemory}. Given that tasks often share sub-procedures, agents could induce one skill from each decomposed subtask rather than from the whole trajectory (\Figure~\ref{fig:intro-illustration}), and early results report better performance. Yet the evidence for this gain stays narrow, covering limited domains, few models, and text skills only. 
% The evidence is also indirect, as prior work reports end-task performance without measuring the quality of the induced skills themselves. 
% \jz{we could also remove this sentence or compress it.}
Answering our central question therefore
% \jz{"answer" unclear (answer to what). Rephrase the sentence.}
requires an analysis that compares task-level with subtask-level induction under identical conditions across various skill formats, domains, and models.
% , and scores each induced skill.

% The controlled comparison and the two headline results.
We conduct a comprehensive and controlled analysis of how agents should induce skills for reliable cross-task transfer.
Our analysis varies two controls, the skill induction level, whether a skill summarizes a whole trajectory or a single subtask, and the skill format, whether a skill is stored as text or as code.
% that compares task-level with subtask-level skill induction and text with code skill formats.
% \jz{maybe frame this to a bit of a summary first, not to tie it with the axes first. For example, we conduct ... analysis for (some high level goal; e.g. when does skill transfer well, how to induce skills for better usability across different tasks).. Our analysis .. different controls, including ... levels (task vs. subtask) and format (text, code; also a bit explaination of what they mean on the very high level briefly)} 
Our experiments cover three long-horizon benchmarks and 11 open-weight and proprietary models. The skill induction level decides whether skills help. Subtask-level skills lift the agent above the no-memory baseline on average, while task-level skills tend to harm the baseline. The skill format then decides how much skills help or hurt, as text skills transfer better than code skills at both skill induction levels (\Section~\ref{sec:results}).

To further understand these findings, 
% explain why subtask-level and text skills transfer better,
% \jz{maybe can rephrase to not focus on the details that much, but to pitch this as a clear second contribution, sth like "to ground our understanding more systematically" (without specifically say subtask and text skills)} 
we examine two complementary properties of the induced skills: \textit{specificity}, which measures how closely a skill matches real tasks, and \textit{abstractness}, which measures how evenly its relevance spreads across tasks. We show that neither property alone predicts task success, but their product does. We define this product as a \textit{skill utility} score. Consistent with our findings, subtask-level and text skills have higher skill utility (\Section~\ref{sec:why}).

% Contributions as practitioner takeaways, answering the opening question.
Our analysis answers the opening question with three takeaways for practitioners. First, decomposing a task into subtasks and inducing one skill per subtask improves cross-task skill transfer. Second, text skills transfer better than code skills at both levels. Finally, skill utility is a lightweight diagnostic tool that only requires skills and task descriptions for practitioners to assess a skill memory.

\providecommand{\cmark}{\checkmark}
\providecommand{\xmark}{$\times$}
\begin{table*}[t]
\centering
\small
\setlength{\tabcolsep}{3.5pt}
\resizebox{\textwidth}{!}{%
\begin{tabular}{l l l l c c c}
\toprule
Work & Skill Induction Level & Skill Format & Domain & Instruction & Demonstration & Abstraction Rules \\
\midrule
AutoManual \citep{chen2024automanual} & Rule & Text & Household, Websites & \cmark & \cmark & \cmark \\
DynaSaur \citep{nguyen2025dynasaur} & Step & Code & Assistance, Math, QA & \cmark & \cmark & \cmark \\
Reflexion \citep{shinn2023reflexion} & Task & Text & QA, Household, Code & \cmark & \cmark & \xmark \\
ExpeL \citep{Zhao_Huang_Xu_Lin_Liu_Huang_2024} & Task & Text & QA, Household, Websites & \cmark & \xmark & \cmark \\
CLIN \citep{majumder2024clin} & Task & Text & Science Simulation, Household & \cmark & \xmark & \cmark \\
AWM \citep{wang2025agent} & Task & Text & Websites & \cmark & \cmark & \cmark \\
Memp \citep{fang2026mempexploringagentprocedural} & Task & Text & Travel, Household & \cmark & \cmark & \xmark \\
Voyager \citep{wang2024voyager} & Task & Code & Game & \cmark & \cmark & \cmark \\
TroVE \citep{wang2024trove} & Task & Code & Math, Table, Vision & \cmark & \cmark & \xmark \\
ASI \citep{wang2025inducing} & Task & Code & Websites & \cmark & \cmark & \cmark \\
SSO \citep{nottingham2024skill} & Subtask & Text & Science Simulation, Game & \cmark & \xmark & \cmark \\
MUSE \citep{yang-etal-2026-towards} & Subtask & Text & Productivity & \cmark & \xmark & \cmark \\
\citet{shen2026structurallyalignedsubtasklevelmemory} & Subtask & Text & Software Engineering & n/r & n/r & n/r \\
\midrule
\textbf{Ours} & Task, Subtask & Text, Code & Assistance, Office, Data Science & \cmark & \cmark & \cmark \\
\bottomrule
\end{tabular}
}
\caption{Survey of skill-induction methods. Skill induction level is the span a skill is induced over, skill format is the representation of a stored skill, and domain lists the evaluation domains in each work's experiments. The last three columns mark whether the induction prompt includes an instruction that states what to extract, a demonstration that shows one worked extraction, and abstraction rules that drop instance-specific detail. The prompt cells are read from each system's released prompt files or paper appendix, and n/r means the prompt is not released.}
\label{tab:survey}
\end{table*}

\section{Related Work}
\label{sec:related}

\SmallHeading{Skills in agents}
A line of work consolidates an agent's experience into an explicit skill memory \citep{sharma-etal-2022-skill, 10.1145/3586183.3606763}, differing along three axes (\Table~\ref{tab:survey}): (\textit{i}) the skill induction level, including a trajectory \citep{wang2024voyager, Zhao_Huang_Xu_Lin_Liu_Huang_2024, shinn2023reflexion, majumder2024clin, wang2025agent}, a step \citep{nguyen2025dynasaur, wang2026reinforcementlearningselfimprovingagent}, a rule \citep{chen2024automanual, fu2024autoguide}, or a subtask \citep{nottingham2024skill, yang-etal-2026-towards, shen2026structurallyalignedsubtasklevelmemory}; (\textit{ii}) the skill format, including text \citep{Zhao_Huang_Xu_Lin_Liu_Huang_2024, zhu2023ghostminecraftgenerallycapable, majumder2024clin, fang2026mempexploringagentprocedural} or code \citep{wang2024trove, wang2025inducing, ellis2021dreamcoder, grand2024lilo, cai2024large, qian-etal-2023-creator, yuan2024craft}; and (\textit{iii}) the induction prompt, with various instructions, demonstrations, and abstraction rules \citep{hu2026memoryageaiagents}.
Others train memory operations or skill-augmented policies with RL \citep{yan-etal-2026-memory, xia2026skillrl, feng2026searljointoptimizationpolicy, li2026memposelfmemorypolicyoptimization}, retrieve episodes without induction \citep{zheng2024synapse, zhang2026memrlselfevolvingagentsruntime, ahmed2026retrievalofthought, yang2024buffer}, or benchmark skill retrieval from large skill libraries \citep{su2026skillretrievalaugmentationagentic, kang2026skillretlargescalebenchmarkskill}.
We vary the level and the format under one shared prompt and skill retriever, isolating each axis.

\SmallHeading{Cross-Task Skill Transfer}
Existing task-level skill transfer shows gains across multiple domains, covering the web \citep{wang2025agent, wang2025inducing, zheng2025skillweaverwebagentsselfimprove, zhou2025proposeragentevaluator}, GUI control \citep{wang2025mobileagente}, embodied worlds \citep{sarch2024vlm}, software \citep{ouyang2026reasoningbank} or research agents \citep{zhou2025mementofinetuningllmagents}.
Despite the breadth of work on task-level skills, the study of subtask-level skill transfer stays limited.
(\textit{i}) Each work evaluates its skills on limited domains \citep{nottingham2024skill, yang-etal-2026-towards, shen2026structurallyalignedsubtasklevelmemory}. 
(\textit{ii}) End-to-end scores hide each skill's contribution to success despite using statistics including reuse counts or library growth \citep{zhong2026skilllearnbenchbenchmarkingcontinuallearning, tan-etal-2025-membench, fang2026mempexploringagentprocedural, yu2025polyskill}.
(\textit{iii}) Most work shows explicitly that skills bring gains, but other work implies the opposite, as irrelevant context degrades models \citep{pmlr-v202-shi23a, yoran2024making, Cuconasu, liu-etal-2024-lost}, misaligned memories propagate errors \citep{xiong-etal-2026-memory, feng-etal-2025-unraveling}, and even relevant memories can fail to propagate \citep{zhong-etal-2023-mquake, feng-etal-2026-tracking}.
We therefore study when skill transfer helps or hurts, isolating each choice and validating a per-skill utility score against success.

\SmallHeading{Subtask decomposition in long-horizon agents}
Prior work breaks a complex question into atomic sub-questions that are easier to solve \citep{zhou2023leasttomost, khot2023decomposed, dua2022successive, prasad-etal-2024-adapt, sun2023adaplanner}. Long-horizon agents adopt the same idea to keep the growing context short and tackle each subtask in long-horizon tasks \citep{hu-etal-2025-hiagent, ye2026agentfold, sun2026scaling}. However, the subtask agents do not transfer skills across tasks. Some works transfer subtask skills but they are limited to one or two domains and few models \citep{nottingham2024skill, yang-etal-2026-towards, shen2026structurallyalignedsubtasklevelmemory}. We instead study cross-task skill transfer at both induction levels, across two skill formats, three domains, and eleven models.

\section{Cross-Task Skill Transfer}
\label{sec:prelim}

\providecommand{\cmark}{\checkmark}
\providecommand{\xmark}{$\times$}

% Pipeline figure (fig:skill-induction-pipeline) belongs to Section 3; declared here so it lands on page 3 top.
\begin{figure*}[t]
\centering
\resizebox{\textwidth}{!}{%
\fontfamily{ppl}\selectfont
\begin{tikzpicture}[
  font=\small,
  >={Stealth[length=2mm]},
  main/.style={rounded corners=5pt, fill=maincyan, text=black, font=\small, minimum width=46mm, minimum height=6.5mm, align=center},
  trace/.style={rounded corners=3pt, fill=tracebg, draw=black!30, font=\footnotesize, align=center, inner sep=3pt, minimum height=13.5mm},
  skill/.style={rounded corners=2pt, fill=customgreen!14, draw=customgreen!75, font=\small, minimum width=8mm, minimum height=5.5mm, align=center},
  task/.style={rounded corners=2pt, fill=maincyan, font=\footnotesize, align=center, minimum height=8mm, minimum width=13mm, inner sep=2pt},
  fmt/.style={rounded corners=2pt, font=\footnotesize, fill=customgreen!14, draw=customgreen!75, inner sep=2.6pt, minimum height=5.4mm},
  edge/.style={draw=black!45, semithick},
  toM/.style={->, draw=customgreen!70, semithick},
  xarr/.style={->, draw=black!45, semithick},
  sarr/.style={-{Stealth[length=2.2mm]}, draw=black!55, semithick},
  glab/.style={font=\normalsize\bfseries, text=black},
]
\def\LC{3.95} \def\SC{12.05}
% ---- panels (rounded box + navy header) ----
\foreach \x/\r/\name in {0.3/7.6/Task-Level Agent, 8.4/15.7/Subtask-Level Agent}{
  \begin{scope}
    \clip[rounded corners=4pt] (\x,1.75) rectangle (\r,5.3);
    \fill[white] (\x,1.75) rectangle (\r,5.3);
    \fill[navy!80] (\x,4.78) rectangle (\r,5.3);
  \end{scope}
  \draw[rounded corners=4pt, draw=black!22, line width=0.7pt] (\x,1.75) rectangle (\r,5.3);
  \node[font=\small, text=white] at ({(\x+\r)/2},5.04) {\name~\figicon[4.5mm]{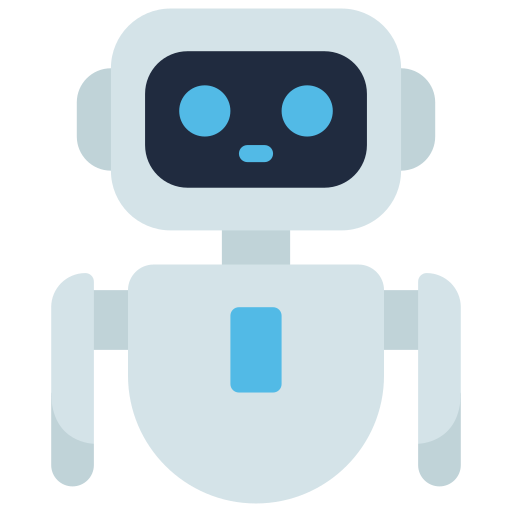}};
}
% ---- induction-level bracket over the panels ----
\draw[black!45] (\LC,5.52) -- (\SC,5.52);
\draw[black!45] (\LC,5.52) -- (\LC,5.34);
\draw[black!45] (\SC,5.52) -- (\SC,5.34);
\node[glab] at ({(\LC+\SC)/2},5.72) {1. Skill Induction Level~\figicon[4mm]{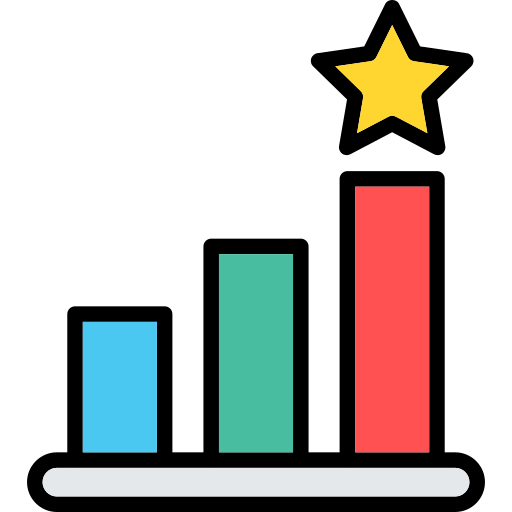}};
% ---- task-level panel content ----
\node[main] (Lmain) at (\LC,4.35) {Current Task $T_i$};
\node[trace, minimum width=54mm] (Ltr) at (\LC,3.2)
  {\textcolor{inkdark}{Task $T_i$}\\[2pt]
   \actpill{$a_1$}\;\obspill{$o_1$}\;\actpill{$a_2$}\;\obspill{$o_2$}\;$\cdots$\\[2pt]
   \textcolor{inkdark}{\scriptsize trajectory $\tau$}};
\node[skill] (Ls) at (\LC,2.12) {$s$};
\node[font=\scriptsize, text=customgreen!60!black] at (\LC+1.75,2.12) {Skill Induction};
\draw[edge] (Lmain) -- (Ltr);
\draw[edge] (Ltr) -- (Ls);
% ---- Subtask panel content ----
\node[main] (Smain) at (\SC,4.35) {Current Task $T_i$};
\node[trace, minimum width=27mm] (Ss1) at (\SC-1.85,3.2)
  {\textcolor{inkdark}{Subtask $T_{i,1}$}\\[2pt]
   \actpill{$a_1$}\;\obspill{$o_1$}\;$\cdots$\\[2pt]
   \textcolor{inkdark}{\scriptsize sub-trajectory $\tau_1$}};
\node[trace, minimum width=27mm] (Ss2) at (\SC+1.85,3.2)
  {\textcolor{inkdark}{Subtask $T_{i,2}$}\\[2pt]
   $\cdots$\\[2pt]
   \textcolor{inkdark}{\scriptsize sub-trajectory $\tau_2$}};
\node[skill] (Ss1s) at (\SC-1.85,2.12) {$s_1$};
\node[skill] (Ss2s) at (\SC+1.85,2.12) {$s_2$};
\node[font=\scriptsize, text=customgreen!60!black] at (\SC,2.12) {Skill Induction};
\draw[edge] (Smain) -- (Ss1);
\draw[edge] (Smain) -- (Ss2);
\draw[edge] (Ss1) -- (Ss1s);
\draw[edge] (Ss2) -- (Ss2s);
\draw[sarr] (Ss1) -- node[above, font=\tiny, text=black!70] {Summary} (Ss2);
% ---- per-agent memory row: earlier -> memory (text/code) -> later, plus retrieve-to-current ----
\node[task, anchor=west, font=\scriptsize] (prevL) at (0.35,0.63) {Earlier\\ Tasks $T_{<i}$};
\node[rounded corners=3pt, fill=customgreen!7, draw=customgreen!45, minimum width=40mm, minimum height=11mm] (ML) at (\LC,0.61) {};
\node[font=\scriptsize, anchor=center] at (\LC,0.9) {Skill Memory $M$~\figicon[4.5mm]{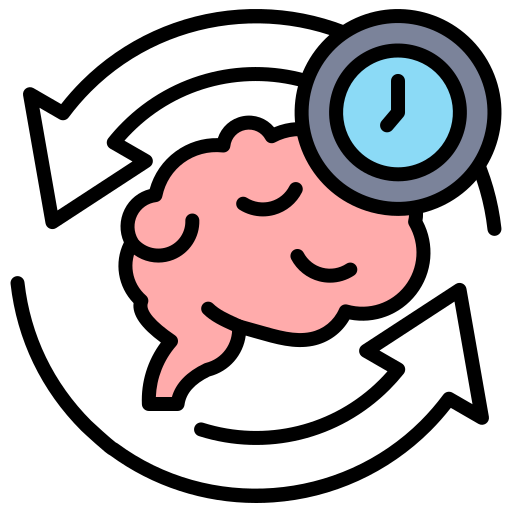}};
\node[fmt, text=fmtorange, font=\scriptsize] at (\LC-0.7,0.38) {Text~\figicon[3mm]{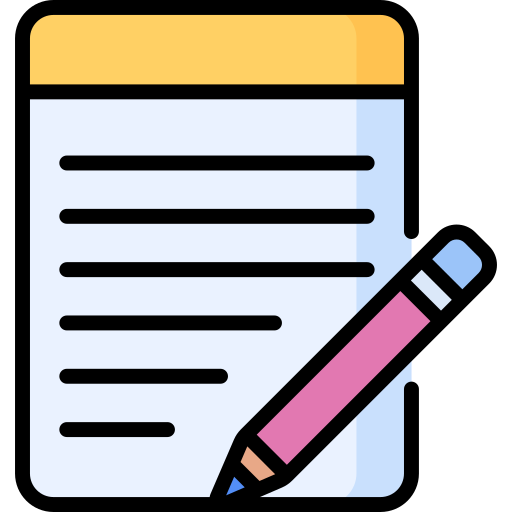}};
\node[fmt, text=fmtblue, font=\scriptsize] at (\LC+0.7,0.38) {Code~\figicon[3mm]{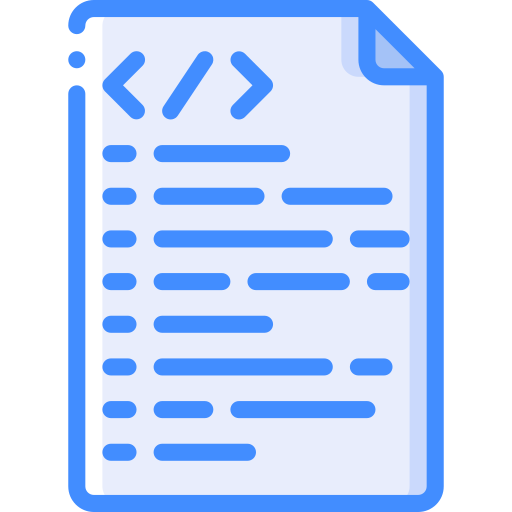}};
\node[task, anchor=east, font=\scriptsize] (latL) at (7.55,0.63) {Later\\ Tasks $T_{>i}$};
\draw[toM] (prevL) -- (prevL-|ML.west);
\draw[->, draw=fmtblue!80, semithick] (ML.east|-latL.west) -- (latL);
\node[font=\scriptsize, text=customgreen!60!black] at (1.1,0.02) {Skill Induction};
\node[font=\scriptsize, text=fmtblue] at (6.8,0.02) {Skill Retrieval};
\draw[-{Stealth[length=2.2mm]}, draw=fmtblue!80, semithick] (1.95,1.05) -- (0.75,1.5) -- (0.75,4.35) -- (Lmain.west);
\node[font=\scriptsize, text=fmtblue, rotate=90] at (0.56,2.75) {Skill Retrieval};
\node[task, anchor=west, font=\scriptsize] (prevS) at (8.45,0.63) {Earlier\\ Tasks $T_{<i}$};
\node[rounded corners=3pt, fill=customgreen!7, draw=customgreen!45, minimum width=40mm, minimum height=11mm] (MS) at (\SC,0.61) {};
\node[font=\scriptsize, anchor=center] at (\SC,0.9) {Skill Memory $M$~\figicon[4.5mm]{figures/memory.png}};
\node[fmt, text=fmtorange, font=\scriptsize] at (\SC-0.7,0.38) {Text~\figicon[3mm]{figures/text.png}};
\node[fmt, text=fmtblue, font=\scriptsize] at (\SC+0.7,0.38) {Code~\figicon[3mm]{figures/code.png}};
\node[task, anchor=east, font=\scriptsize] (latS) at (15.65,0.63) {Later\\ Tasks $T_{>i}$};
\draw[toM] (prevS) -- (prevS-|MS.west);
\draw[->, draw=fmtblue!80, semithick] (MS.east|-latS.west) -- (latS);
\node[font=\scriptsize, text=customgreen!60!black] at (9.2,0.02) {Skill Induction};
\node[font=\scriptsize, text=fmtblue] at (14.9,0.02) {Skill Retrieval};
\draw[-{Stealth[length=2.2mm]}, draw=fmtblue!80, semithick] (10.05,1.05) -- (8.7,1.5) -- (8.7,4.35) -- (Smain.west);
\node[font=\scriptsize, text=fmtblue, rotate=90] at (8.55,2.75) {Skill Retrieval};
% ---- skill-format bracket under the four format chips ----
\draw[black!45] (\LC-0.7,-0.28) -- (\SC+0.7,-0.28);
\draw[black!45] (\LC-0.7,-0.28) -- (\LC-0.7,-0.1);
\draw[black!45] (\LC+0.7,-0.28) -- (\LC+0.7,-0.1);
\draw[black!45] (\SC-0.7,-0.28) -- (\SC-0.7,-0.1);
\draw[black!45] (\SC+0.7,-0.28) -- (\SC+0.7,-0.1);
\node[glab] (fl) at (8.0,-0.52) {2. Skill Format~\figicon[4.5mm]{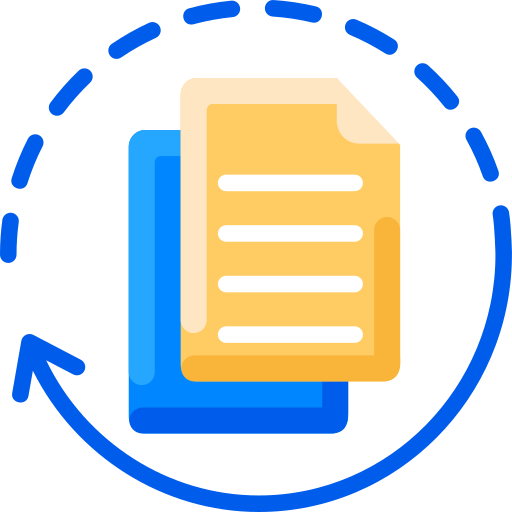}};
\draw[toM] (Ls)  -- (Ls|-ML.north);
\draw[toM] (Ss1s) -- (Ss1s|-MS.north);
\draw[toM] (Ss2s) -- (Ss2s|-MS.north);
\end{tikzpicture}%
}
\caption{Illustration of two skill induction levels and two skill formats. (\textit{i}) The task-level agent runs the current task as one trajectory of actions and observations and induces one skill from the whole trajectory. The subtask-level agent decomposes the task into subtasks, passes a running summary between consecutive subtasks, and induces one skill from each sub-trajectory. The two agents differ only in the skill induction level. (\textit{ii}) Each agent stores its induced skills in its own memory as text notes or code functions and retrieves relevant skills back into its context, so skills induced on earlier tasks serve later ones. Crossing the two levels with the two formats and a no-memory baseline gives the six conditions we compare.}
\label{fig:skill-induction-pipeline}
\end{figure*}

% The section's purpose and roadmap in one paragraph.
This section formalizes cross-task skill transfer, where an LLM agent solves a stream of tasks and its experience on earlier tasks affects how it solves later ones. We introduce two agents that map to two skill induction levels (\Section~\ref{sec:setup}), a skill memory that induces, stores, and retrieves skills in two different formats (\Section~\ref{sec:skillmem}), and a definition of cross-task skill transfer (\Section~\ref{sec:transfer}). \Figure~\ref{fig:skill-induction-pipeline} shows the illustration, and the full prompts are in \Appendix~\ref{app:prompts}.

\subsection{Agents}
\label{sec:setup}

% The shared POMDP substrate: per-step interaction and final-state reward.
An agent solves each task in a partially observable environment \citep{KAELBLING199899}. A task specifies a natural-language goal and an initial environment state. At step $t$, an LLM policy $\pi$ generates an action $a_t$ based on the interaction so far, and the environment returns an observation $o_t$. When the agent signals completion or reaches a step limit, an evaluator gives the final state a reward $r \in [0,1]$.

% The task-level agent keeps one flat, growing history.
\SmallHeading{Task-level agent}
The task-level agent is a flat ReAct loop \citep{yao2023react}. It appends ($\oplus$) every action and observation into one context $h_t = h_{t-1} \oplus (a_t, o_t)$ that grows from an initial prompt $h_0$ encoding the goal, and draws each action $a_t$ from $\pi(\cdot \mid h_{t-1})$. The loop ends in the whole-task trajectory $\tau = h_0 \oplus (a_1, o_1) \oplus \dots \oplus (a_T, o_T)$.

% The subtask-level agent runs a plan-execute-summarize cycle, carrying only a summary across subtasks.
\SmallHeading{Subtask-level agent} 
The subtask-level agent runs the same environment and policy, but it decomposes each task into subtasks \citep{zhou2023leasttomost, prasad-etal-2024-adapt} with a cycle of three roles. A planner reads the goal and the running summary, then proposes the next subtask or declares the task complete. An executor runs a ReAct loop on the current subtask, producing a sub-trajectory $\tau_k$ for subtask $k$. A summarizer then compresses $\tau_k$ into the running summary for the next cycle \citep{hu-etal-2025-hiagent}. The cycle repeats until the planner declares completion or a subtask limit is reached, ending in the task trajectory $\tau = (\tau_1, \dots, \tau_K)$. The task-level agent is thus a special case of the subtask-level agent, where the whole task forms the single subtask and only the executor runs.

\subsection{Skill Memory}
\label{sec:skillmem}

% Introduce the memory and the two operators.
A skill memory $M$ stores skills from the agent's experience for reuse on later tasks. An induction operator turns the trajectory of a completed task or subtask into a skill, and a retrieval operator reads relevant skills from $M$ before a task or subtask.

% Skill = description + body; the two formats differ in the body.
\SmallHeading{Skill format}
Each skill has a short natural-language description, and a body that carries the main content in text or code format inspired by \citet{wang2025agent, wang2025inducing}. A text skill writes the body as a workflow note, listing the procedure and its environment-specific caveats. A code skill writes the body as a Python function with instance-specific values as parameters and environmental caveats as code comments.

% How induction writes a skill, and that its span is the agent's unit.
\SmallHeading{Skill induction}
Skills are induced at two levels. Task-level induction turns the completed task trajectory $\tau$ into one skill, and subtask-level induction turns each completed sub-trajectory $\tau_k$ into one skill.
This one-skill-per-trajectory rule prevents task-level induction from splitting a trajectory into several subtask-like skills, which would blur the two levels.
The two levels share the same induction prompt, so the contrast isolates only the skill induction level.
Examples are in \Appendix~\ref{app:skill-examples}.
% \nb{It is ambiguous as to whether the task-level skill induction produces one skill per task or if it produces multiple skills. It is better to state this more directly. The figure seems to suggest that there could be multiple skills induced per task ($s_1$, and $s_2$ boxes). It is better to expand this skill induction paragraph, since this is one of the key foci of your paper. It is also useful for the readers to refer to some actual examples of these skills if you have them in the appendix or elsewhere in the paper.}

% How retrieval selects and injects skills, per agent and per format.
\SmallHeading{Skill retrieval}
The retrieval operator $R(M, q)$ embeds skill descriptions and the query $q$ with an embedding model{\interfootnotelinepenalty10000\footnote{We use \texttt{all-MiniLM-L6-v2}, \url{https://huggingface.co/sentence-transformers/all-MiniLM-L6-v2}.}} and injects the top matches into the agent's context. Text and code skills are retrieved in the same way by matching the description only. The task-level agent queries once with the task instruction, and the subtask-level agent queries at each subtask with the task instruction or the subtask text. 
In our setup, both agents retrieve the same number of skills per query (top 5 skills).
Every retrieved skill enters the context as its description and body, and a code skill is also loaded into the namespace so the agent can call it by name. The memory also prunes code skills that fail to load and merges or drops near-duplicate descriptions (\Appendix~\ref{app:memory}). 
We verify in \Appendix~\ref{app:selfretr} that the differences among skill conditions come from the skills themselves rather than from retrieval quality.
% We validate the retrieval embedder in \Appendix~\ref{app:selfretr} so that retrieval quality does not ...

\subsection{Cross-Task Skill Transfer}
\label{sec:transfer}

% Combine agents and memory into the formal definition of the phenomenon.
An agent from \Section~\ref{sec:setup} solves a stream of tasks $T_1, \dots, T_N$ against the shared memory of \Section~\ref{sec:skillmem}. It writes one skill after each task or subtask and reads them before each, so a skill induced on an earlier task can be retrieved on a later one. We call such skill reuse \textit{cross-task skill transfer}.

% Name the two design choices the study varies, one from the agents and one from the memory.
\SmallHeading{Design choices}
Our study varies two choices.
(\textit{i}) The skill induction level is the span of task trajectory that each skill is induced from, the whole task for the task-level agent or a single subtask for the subtask-level agent.
(\textit{ii}) The skill format is how a skill is stored, either a text note, a code function, or none when the memory is off.
% Only design choices are varied
Note that the two agents differ only in the skill induction level, plus the subtask-level agent's added planner and summarizer prompts. Every other aspect of the prompts, memory, and retrieval is identical.
The remaining difference cancels in our comparisons, as each agent with skills is measured against the same agent without skills, so the results reflect the effect of the skills induced at each level rather than the agents themselves.

% Experiment Setup
\section{Experiment Setup}
\label{sec:experiments}

\SmallHeading{Benchmarks}
We use three long-horizon benchmarks that span diverse domains. \ul{AppWorld} \citep{trivedi-etal-2024-appworld} covers multi-app tool use in a Python REPL over nine simulated apps with documented APIs. \ul{OfficeBench} \citep{wang2024officebench} covers office-document workflows over apps such as Email, Word, Excel, and PDF. \ul{KramaBench} \citep{lai2025kramabench} covers data-science pipelines over real files from six scientific domains. We evaluate on the AppWorld test-challenge split (417 tasks), 300 OfficeBench tasks, and the 92 deterministically graded KramaBench tasks.

\SmallHeading{Models}
We evaluate three Mixture-of-Experts (MoE) models (\ul{Qwen3-235B-A22B} \cite{yang2025qwen3}, \ul{GPT-OSS-120B} \cite{agarwal2025gpt}, and \ul{Nemotron-Super-120B} \cite{nvidia2026nemotron3superopen}), dense models (\ul{Qwen3 at 4B, 8B, 14B, 32B} \cite{yang2025qwen3} and \ul{Gemma-3 at 4B, 12B, 27B} \cite{gemmateam2025gemma3technicalreport}), and one commercial model (\ul{Gemini-3.1-Pro}{\interfootnotelinepenalty10000\footnote{\url{https://ai.google.dev/gemini-api/docs/models/gemini-3.1-pro-preview}}}). We serve the MoE models on Amazon Bedrock, Gemini-3.1-Pro on Google Cloud, and each dense model with vLLM on one GH200. We keep each provider's default sampling, and give every model the same prompts.
\Appendix~\ref{app:eval} reports the reasoning and precision settings, the latency and tokens of each model, and the total cost.
We also rerun three MoE models and Qwen3-32B with two reduced induction prompts and a reshuffled task order to measure run-to-run variance.

\SmallHeading{Comparison Conditions}
We cross the two axes of \Section~\ref{sec:prelim}, skill induction level (Task-level or Subtask-level) and skill format (None, Text, or Code), into six conditions: \ul{Task}, \ul{Task+Text}, \ul{Task+Code}, \ul{Subtask}, \ul{Subtask+Text}, and \ul{Subtask+Code}, where Task and Subtask alone carry no memory. The ``+Text'' tag adds a text skill and the ``+Code'' tag adds a code skill.
% The induction prompt is identical across the two levels, so each contrast isolates one axis. 
% We compare each agent with skills against the same agent without skills, so differences between the two agents cancel out and the comparison reflects only the effect of the skills induced at each level. (mentioned in sec 3.3)
We limit the task-level agent at 50 ReAct steps and the subtask-level agent at 15 subtasks under a shared 50-step executor budget.\footnotemark

\SmallHeading{Evaluation}
% We evaluate along two axes, performance and efficiency. 
For performance, we report \ul{task success}, the average score in $[0,1]$ from each benchmark's official evaluator. For efficiency, we report \ul{latency}, the wall-clock time per task, and \ul{dependency}, a measure of the compute spent on the growing context \citep{zhou2026mem} (\Appendix~\ref{app:eval}).

\section{Skill Induction Level and Format Impact Whether a Skill Helps}
\label{sec:results}

We now test how each of the two induction choices affects cross-task skill transfer, varying the skill induction level under both skill formats (\Section~\ref{sec:results:gran}) and the format at both levels (\Section~\ref{sec:results:fmt}).

\begin{boxblue}
\textbf{Takeaway.}
Skills help often when they are induced at the subtask level, while task-level induction often makes the same memory harmful. Text skills mostly help more than code skills at both induction levels.
\end{boxblue}

\begin{table*}[t!]
\centering
\scriptsize \setlength{\tabcolsep}{3pt} \renewcommand{\arraystretch}{0.95}
\resizebox{\textwidth}{!}{%
\begin{tabular}{ll ccc ccc}
\toprule
 & & \multicolumn{3}{c}{Task-Level} & \multicolumn{3}{c}{Subtask-Level} \\
\cmidrule(lr){3-5}\cmidrule(lr){6-8}
Benchmark & Model & None & +Text & +Code & None & +Text & +Code \\
\midrule
\multirow{10}{*}{AppWorld}
 & Qwen3-235B-A22B     & 7.4\ci{5.0}{10.1} & 18.0\ci{14.4}{21.8} & 14.4\ci{11.0}{18.0} & 27.3\ci{23.0}{31.7} & 35.5\ci{30.9}{40.0} & \textbf{35.7}\ci{31.2}{40.3} \\
 & GPT-OSS-120B        & \textbf{27.3}\ci{23.3}{31.7} & 17.0\ci{13.4}{20.6} & 1.0\ci{0.2}{1.9} & 26.4\ci{22.3}{30.5} & 25.2\ci{21.1}{29.5} & 23.7\ci{19.7}{27.8} \\
 & Nemotron-Super-120B & 8.9\ci{6.2}{11.8} & 4.3\ci{2.4}{6.2} & 1.4\ci{0.5}{2.6} & 17.7\ci{14.1}{21.6} & \textbf{21.6}\ci{17.7}{25.4} & 18.5\ci{14.9}{22.3} \\
 & Qwen3-4B            & 0.0\ci{0.0}{0.0} & 0.2\ci{0.0}{0.7} & 0.0\ci{0.0}{0.0} & 0.2\ci{0.0}{0.7} & 0.7\ci{0.0}{1.7} & \textbf{1.4}\ci{0.5}{2.6} \\
 & Qwen3-8B            & 0.2\ci{0.0}{0.7} & 0.0\ci{0.0}{0.0} & 0.0\ci{0.0}{0.0} & 2.2\ci{1.0}{3.6} & \textbf{3.1}\ci{1.7}{5.0} & 1.9\ci{0.7}{3.4} \\
 & Qwen3-14B           & 0.5\ci{0.0}{1.2} & 0.7\ci{0.0}{1.7} & 0.2\ci{0.0}{0.7} & 6.7\ci{4.3}{9.4} & 7.2\ci{4.8}{9.8} & \textbf{8.4}\ci{5.8}{11.0} \\
 & Qwen3-32B           & 1.9\ci{0.7}{3.4} & 2.9\ci{1.4}{4.6} & 1.4\ci{0.5}{2.6} & 7.7\ci{5.3}{10.3} & \textbf{10.3}\ci{7.4}{13.2} & 7.2\ci{4.8}{9.8} \\
 % & Gemma-3-4B          & \textbf{0.0}\ci{0.0}{0.0} & \textbf{0.0}\ci{0.0}{0.0} & \textbf{0.0}\ci{0.0}{0.0} & \textbf{0.0}\ci{0.0}{0.0} & \textbf{0.0}\ci{0.0}{0.0} & \textbf{0.0}\ci{0.0}{0.0} \\
 % & Gemma-3-12B         & 0.0\ci{0.0}{0.0} & 0.2\ci{0.0}{0.7} & 0.2\ci{0.0}{0.7} & \textbf{1.7}\ci{0.5}{3.1} & 1.4\ci{0.5}{2.6} & 1.0\ci{0.2}{1.9} \\
 & Gemma-3-27B         & 0.5\ci{0.0}{1.2} & 0.5\ci{0.0}{1.2} & 0.0\ci{0.0}{0.0} & 4.6\ci{2.6}{6.7} & 4.3\ci{2.4}{6.5} & \textbf{4.8}\ci{2.9}{7.0} \\
 & Gemini-3.1-Pro      & 68.1\ci{63.5}{72.4} & 58.0\ci{53.2}{62.6} & 52.3\ci{47.5}{57.1} & 68.3\ci{63.8}{72.7} & 72.4\ci{68.1}{76.7} & \textbf{77.5}\ci{73.4}{81.5} \\
\cmidrule(lr){2-8}
 & Average                & 10.4\ci{9.6}{11.3} & 9.3\ci{8.4}{10.2} & 6.5\ci{5.8}{7.1} & 14.8\ci{13.5}{16.2} & \textbf{16.5}\ci{15.1}{18.0} & 16.4\ci{15.1}{17.7} \\
\cmidrule(lr){1-8}
\multirow{10}{*}{OfficeBench}
 & Qwen3-235B-A22B     & \textbf{43.0}\ci{37.3}{48.7} & 38.3\ci{33.0}{44.0} & 36.7\ci{31.3}{42.0} & 38.7\ci{33.3}{44.3} & 41.3\ci{35.7}{47.0} & 38.7\ci{33.3}{44.0} \\
 & GPT-OSS-120B        & 32.3\ci{27.0}{37.7} & 29.3\ci{24.3}{34.3} & 5.0\ci{2.7}{7.7} & 38.0\ci{32.7}{43.3} & \textbf{42.0}\ci{36.7}{47.7} & 40.7\ci{35.3}{46.3} \\
 & Nemotron-Super-120B & 28.3\ci{23.3}{33.7} & 31.7\ci{26.3}{37.0} & 12.7\ci{9.0}{16.7} & 27.3\ci{22.3}{32.3} & \textbf{37.7}\ci{32.3}{43.3} & 25.0\ci{20.0}{30.0} \\
 & Qwen3-4B            & 17.7\ci{13.3}{22.0} & 16.3\ci{12.3}{20.7} & 13.0\ci{9.3}{17.0} & 18.0\ci{13.7}{22.7} & \textbf{25.3}\ci{20.7}{30.3} & 20.3\ci{16.0}{25.0} \\
 & Qwen3-8B            & \textbf{25.0}\ci{20.0}{30.0} & 15.7\ci{11.7}{20.0} & 16.3\ci{12.3}{20.7} & 19.0\ci{14.7}{23.7} & \textbf{25.0}\ci{20.3}{30.0} & 24.7\ci{19.7}{29.7} \\
 & Qwen3-14B           & 28.3\ci{23.3}{33.7} & 24.0\ci{19.3}{29.0} & 27.3\ci{22.3}{32.3} & 27.0\ci{22.0}{32.0} & \textbf{32.0}\ci{27.0}{37.3} & 30.3\ci{25.3}{35.7} \\
 & Qwen3-32B           & 34.3\ci{29.0}{39.7} & 32.7\ci{27.3}{38.0} & 35.3\ci{30.0}{41.0} & 33.3\ci{28.3}{38.7} & \textbf{36.0}\ci{30.7}{41.7} & 31.7\ci{26.7}{37.0} \\
 % & Gemma-3-4B          & \textbf{3.7}\ci{1.7}{6.0} & 3.0\ci{1.3}{5.0} & 2.3\ci{0.7}{4.3} & 2.7\ci{1.0}{4.7} & 1.0\ci{0.0}{2.3} & 0.7\ci{0.0}{1.7} \\
 % & Gemma-3-12B         & 10.0\ci{6.7}{13.7} & 19.7\ci{15.3}{24.3} & 11.0\ci{7.7}{14.7} & 13.7\ci{10.0}{17.7} & \textbf{21.3}\ci{17.0}{26.0} & 20.0\ci{15.7}{24.7} \\
 & Gemma-3-27B         & \textbf{28.3}\ci{23.3}{33.7} & 26.7\ci{21.7}{31.7} & 14.7\ci{10.7}{18.7} & 24.0\ci{19.0}{29.0} & 19.0\ci{14.7}{23.7} & 23.7\ci{19.0}{28.7} \\
 & Gemini-3.1-Pro      & 46.3\ci{40.7}{52.0} & 41.0\ci{35.3}{46.7} & 41.7\ci{36.3}{47.3} & 47.3\ci{41.7}{53.0} & 46.0\ci{40.3}{51.7} & \textbf{48.7}\ci{43.0}{54.3} \\
\cmidrule(lr){2-8}
 & Average                & 27.0\ci{23.6}{30.4} & 25.3\ci{22.1}{28.8} & 19.6\ci{17.0}{22.4} & 26.3\ci{23.0}{29.7} & \textbf{29.7}\ci{26.2}{33.2} & 27.7\ci{24.2}{31.2} \\
\cmidrule(lr){1-8}
\multirow{10}{*}{KramaBench}
 & Qwen3-235B-A22B     & 52.8\ci{43.0}{62.5} & 49.3\ci{39.2}{59.2} & 51.5\ci{41.1}{61.2} & 52.4\ci{42.5}{62.2} & \textbf{55.3}\ci{45.4}{64.7} & 52.7\ci{42.9}{62.7} \\
 & GPT-OSS-120B        & 31.7\ci{22.8}{41.1} & 28.4\ci{19.6}{37.4} & 22.5\ci{14.2}{31.1} & 48.1\ci{38.3}{58.2} & 49.8\ci{39.6}{59.5} & \textbf{50.9}\ci{40.8}{60.6} \\
 & Nemotron-Super-120B & 53.3\ci{43.2}{63.2} & 52.1\ci{42.1}{62.2} & 43.3\ci{33.5}{53.4} & \textbf{59.8}\ci{50.0}{69.1} & 57.6\ci{47.6}{67.6} & 49.9\ci{40.1}{59.7} \\
 & Qwen3-4B            & 8.3\ci{3.3}{14.4} & 12.3\ci{6.2}{19.0} & 13.3\ci{6.9}{20.4} & 13.8\ci{7.5}{20.8} & \textbf{15.1}\ci{8.3}{22.5} & 11.5\ci{5.5}{18.1} \\
 & Qwen3-8B            & 10.7\ci{4.8}{17.2} & \textbf{15.5}\ci{8.7}{22.9} & 14.9\ci{8.3}{22.0} & 14.4\ci{7.7}{21.5} & 11.2\ci{5.4}{17.7} & 14.1\ci{7.5}{21.5} \\
 & Qwen3-14B           & 20.1\ci{12.5}{28.2} & 14.9\ci{8.5}{22.0} & 23.1\ci{14.7}{31.6} & 19.1\ci{11.5}{27.1} & \textbf{23.5}\ci{15.2}{32.5} & 20.3\ci{12.5}{28.6} \\
 & Qwen3-32B           & 30.0\ci{21.1}{39.2} & 25.8\ci{17.6}{34.8} & 30.2\ci{21.4}{39.3} & 34.3\ci{25.0}{43.9} & \textbf{39.0}\ci{29.3}{48.6} & 38.9\ci{29.2}{48.7} \\
 % & Gemma-3-4B          & 1.5\ci{0.0}{4.0} & 1.1\ci{0.0}{3.3} & 1.9\ci{0.0}{4.9} & \textbf{6.0}\ci{1.6}{11.4} & 4.3\ci{1.0}{8.6} & 2.2\ci{0.0}{5.4} \\
 % & Gemma-3-12B         & 16.3\ci{9.7}{23.7} & 14.0\ci{7.3}{21.2} & 11.8\ci{6.0}{18.2} & \textbf{17.9}\ci{10.7}{25.7} & 17.7\ci{10.6}{25.3} & 14.2\ci{7.9}{21.3} \\
 & Gemma-3-27B         & 19.7\ci{12.2}{27.7} & 22.0\ci{14.0}{30.6} & 18.6\ci{11.3}{26.5} & \textbf{25.3}\ci{16.7}{34.4} & 23.9\ci{15.9}{32.7} & 24.2\ci{16.0}{33.1} \\
 & Gemini-3.1-Pro      & 74.3\ci{65.6}{82.5} & 74.1\ci{65.3}{82.4} & 75.1\ci{66.4}{83.3} & \textbf{75.2}\ci{66.5}{83.2} & 73.7\ci{64.9}{82.0} & 72.4\ci{63.6}{80.8} \\
\cmidrule(lr){2-8}
 & Average                & 29.0\ci{24.1}{34.0} & 28.1\ci{23.2}{33.3} & 27.8\ci{23.0}{32.9} & 33.3\ci{28.0}{38.7} & \textbf{33.7}\ci{28.5}{39.1} & 31.9\ci{26.8}{37.1} \\
\midrule
\multicolumn{2}{l}{Average (11 models)} & 22.1\ci{20.2}{24.2} & 20.9\ci{18.9}{22.9} & 18.0\ci{16.1}{19.9} & 24.8\ci{22.7}{27.0} & \textbf{26.7}\ci{24.5}{28.8} & 25.3\ci{23.3}{27.4} \\
\bottomrule
\end{tabular}}
\caption{Task success (\%) of the task-level and subtask-level agents, without (None) or with induced skills as text notes (+Text) or code functions (+Code), on AppWorld, OfficeBench, and KramaBench. We show nine of the eleven models and defer the two weakest, Gemma-3-4B and Gemma-3-12B, to \Table~\ref{tab:main-full}. Every Average row still covers all eleven models. Brackets are 95\% task-bootstrap confidence intervals. On most models and benchmarks the highest success falls in a Subtask-level column, and adding skills mostly raises the success of the subtask-level agent while it lowers that of the task-level agent.}
% \nb{Do we need the 4B, 8B, and 12B models? The performance numbers are quite low on these tasks. We could move them to the appendix. These table numbers are too small. }}
\label{tab:main}
\end{table*}

\subsection{Skills Help at the Subtask Level}
\label{sec:results:gran}

% Intro: question + roadmap.
We first ask whether induced skills help the agents later, and how the answer depends on the skill induction level. Additional results are in \Appendix~\ref{app:full-table}.

\begin{figure}[t]
\centering
\includegraphics[width=\linewidth]{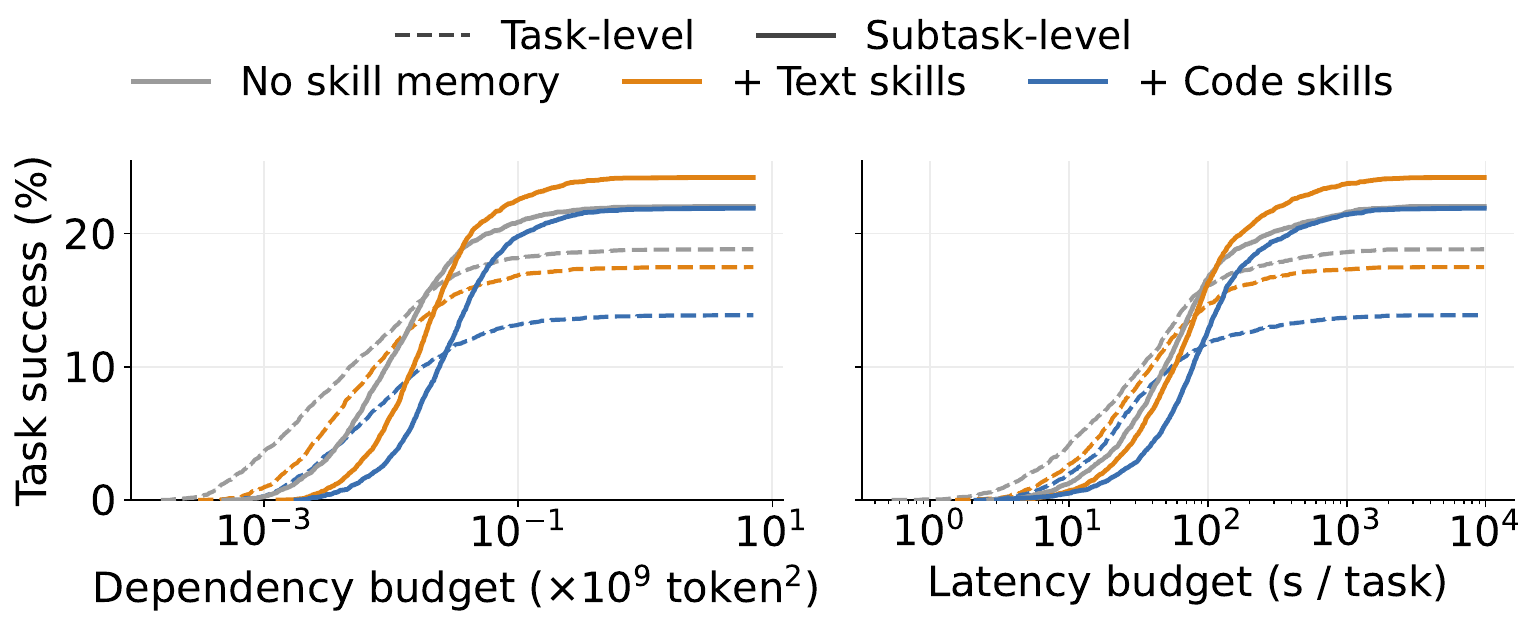}
\caption{Task success rate within a per-task budget of dependency, i.e. the compute spent (left), and latency, i.e., the wall-clock time (right) per task (\Section~\ref{sec:experiments}). We average over models and benchmarks for each skill format and skill induction level. The task-level agent wins only at the smallest budgets, and from moderate budgets on the subtask-level agent with skills solves more tasks at equal cost.}
\label{fig:efficiency}
\end{figure}

% % Headline results: gap + interaction + sign split + scaling.
% \SmallHeading{Performance (task success)} \Table~\ref{tab:main} presents task success across the six conditions.
% % 1. Subtask+skill beats Task+skill.
% Under the same skill format, the subtask-level agent beats the task-level agent for almost every model on every benchmark, by $6.3$ points with Text skills and $6.7$ with Code averaged over all models.
% % 2. Bracket out the no-memory gap, then the skill effect flips sign with the induction level.
% The gap does not come from the agent itself, as the no-memory subtask-level agent only beats its task-level counterpart by $3.2$ points. The equipped skills contribute to the rest of the gap. Under the same format and the same induction prompt, skills induced from subtasks raise success by up to $3$ points over the no-memory baseline, while skills induced from whole tasks lower it by up to $7.4$ points, so skills mostly help at the subtask level.
% % Supplementary check: swap the memories, the gain follows the skills (details in the appendix).
% To further check that the gap is not only the agent's effect, we equip the task-level agent with the subtask-level skills, which beat its own task-level skills on every benchmark (\Appendix~\ref{app:swap}).
% % Ranking (brief pointer)
% Our conclusions also hold under various induction prompts (\Figure~\ref{fig:prompt-ablation}).

% Headline results: within-agent skill effect, sign flips with induction level.
\SmallHeading{Performance (task success)} \Table~\ref{tab:main} presents task success across the six conditions.
\footnotetext{The planner prompt requires an agent to generate at most 5 subtasks, which only serves as a soft control and the actual limit is 15.}
% 1. The comparison is each agent WITH skills vs the SAME agent WITHOUT skills.
% For each task-level or subtask-level agent, we compare it with and without the induced memory as the effect of skills.
% 2. The skill effect flips sign with the induction level.
Skills induced from whole tasks lower the task-level agent's average success, by $1.2$ points with Text skills and $4.1$ with Code. The same induction prompt applied at the subtask level instead raises average success, by $1.9$ points with Text and $0.5$ with Code.
% 3. Scope: consistent for Text, on balance for Code.
The subtask-level gain is consistent for Text skills, which help on all three benchmarks, while Code skills help on two of the three and lose $1.4$ points on KramaBench.
We analyze the cases outside these trends in \Appendix~\ref{app:exceptions}.
% 4. Causal check: swap the memories, the effect follows the skills.
To confirm that the effect travels with the skills rather than the agent, we equip the task-level agent with the subtask-level skills, which beat its own task-level skills on every benchmark. 
The effect is also not correlated by the outcomes of the source tasks, as the two levels induce from solved and unsolved tasks at close rates.
% Robustness pointer.
Our conclusions also hold under runs with various induction prompts and a reshuffled task order (\Appendix~\ref{app:swap}, and \ref{app:shuffle}).

\SmallHeading{Efficiency (latency and dependency)} We next examine task success under a per-task budget of latency and dependency, which measure the time and the compute spent on the growing context respectively (\Section~\ref{sec:experiments}).
% 1. How to read the figure.
\Figure~\ref{fig:efficiency} plots task success rate within each budget, so a vertical slice compares conditions at equal cost.
% 2. Small budgets favor the task-level agent; from moderate budgets on, Subtask wins under every format.
Small budgets favor the task-level agent. From moderate budgets on, the subtask-level agent overtakes it under every format and saturates higher, so the subtask-level advantage is not bought by extra cost. The same crossover appears on every benchmark (\Figure~\ref{fig:budget-bench}).
% 3. Within-agent reading at equal cost mirrors the sign flip.
% The within-agent contrast also survives cost matching. 
At equal cost the subtask-level agent with skills stays above its no-memory curve, while the task-level agent with skills stays below its own.

% \SmallHeading{Varying task difficulty} The comparison reaches the same conclusion at every task difficulty.
% % 1. Difficulty: same ranking in every official stratum.
% We split tasks into easy, medium, and hard using each benchmark's official labels, including the annotated difficulty on AppWorld, the number of apps on OfficeBench, and the easy or hard tag on KramaBench. \Figure~\ref{fig:difficulty} demonstrates the performance comparison within each difficulty stratum. In every stratum and under both formats, the subtask-level agent with skills stays above the task-level agent with skills. In nearly every stratum, the skills also bring a larger gain to the subtask-level agent than to the task-level agent.

\begin{figure*}[t]
\centering
\includegraphics[width=\textwidth]{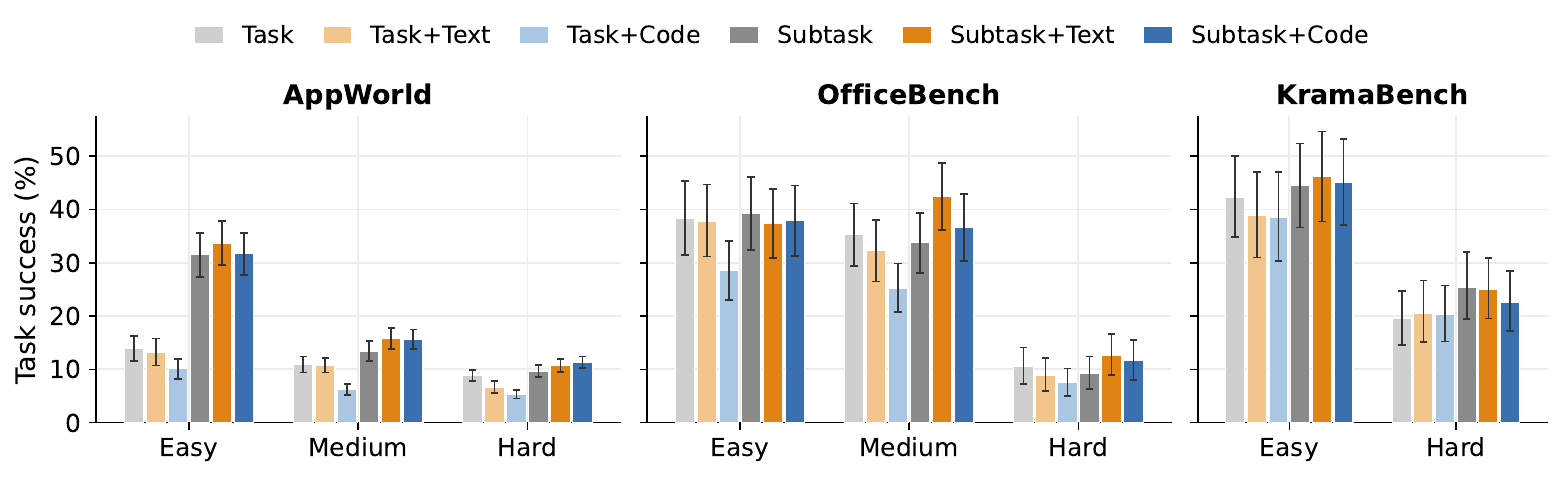}
\caption{Task success of the six conditions by task difficulty, pooled over models with 95\% task-bootstrap confidence intervals. In every stratum and under both formats, the subtask-level agent with skills stays above the task-level agent with skills. In nearly every stratum, the skills bring a larger gain to the subtask-level agent than to the task-level agent, and Text stays at or above Code within each induction level.}
\label{fig:difficulty}
\end{figure*}

\SmallHeading{Varying task difficulty}
% 1. Strata from official labels.
We split tasks into easy, medium, and hard using each benchmark's official labels, including the annotated difficulty on AppWorld and Kramabench and the number of apps on OfficeBench (\Figure~\ref{fig:difficulty}). 
% \Figure~\ref{fig:difficulty} compares all six conditions within each stratum.
% 2. The sign flip persists across strata.
Task-level skills lower success in nearly every stratum under both formats, subtask-level skills raise it in most strata, and in nearly every stratum skills bring a larger gain at the subtask level than at the task level.

\subsection{Text Skills Transfer Better than Code}
\label{sec:results:fmt}

% Intro: question.
Next, we vary the skill format under various skill induction levels.

\SmallHeading{Performance and efficiency}
% 1. Text beats Code at both induction levels.
At the same skill induction level, retrieving text skills beats retrieving code skills, by $2.9$ points on the task-level agent and $1.4$ points on the subtask-level agent averaged over all models (\Table~\ref{tab:main}).
% 2. Same ordering stated as skill effects.
Relative to each agent's no-memory baseline, text skills damage the task-level agent less than code skills, $1.2$ against $4.1$ points, and help the subtask-level agent more, $1.9$ against $0.5$ points.
% 3. Ranking holds at matched cost.
The ranking also survives under various per-task efficiency budgets, as the Text curve stays at or above the Code curve within each induction level at every budget  (\Figure~\ref{fig:efficiency}).

\SmallHeading{Varying task difficulty}
% Positive gap in nearly every stratum.
Across all difficulty strata of \Section~\ref{sec:results:gran}, the Text conditions stay at or above the Code conditions within each skill induction level in nearly every stratum, so the format ranking also holds across difficulty (\Figure~\ref{fig:difficulty}).

% \section{Why Induction Level and Format Matter}
\section{When Do Skills Transfer?}
\label{sec:why}

% Intro: from what wins to why; roadmap.
To understand when certain skills transfer better than others,
% To understand why the subtask level and the text format transfer better, 
we turn our focus from the agents to the skills they induce. We define a per-skill \textit{utility score} computed from the induced skills and the task descriptions alone (\Section~\ref{sec:why:score}), show that it explains why subtask-level and text skills transfer better (\Section~\ref{sec:why:explain}), and check that it agrees with the reuse observed during the task stream (\Section~\ref{sec:why:reuse}).

\begin{boxblue}
\textbf{Takeaway.}
A skill is useful when it is both specific to real tasks and abstract enough to remain relevant across many of them. Neither dimension alone predicts task success. We therefore propose a skill utility score that requires both jointly. Computed directly from the induced skills, it predicts task success and matches observed reuse.
\end{boxblue}

\begin{figure*}[t]
\centering
\includegraphics[width=\textwidth]{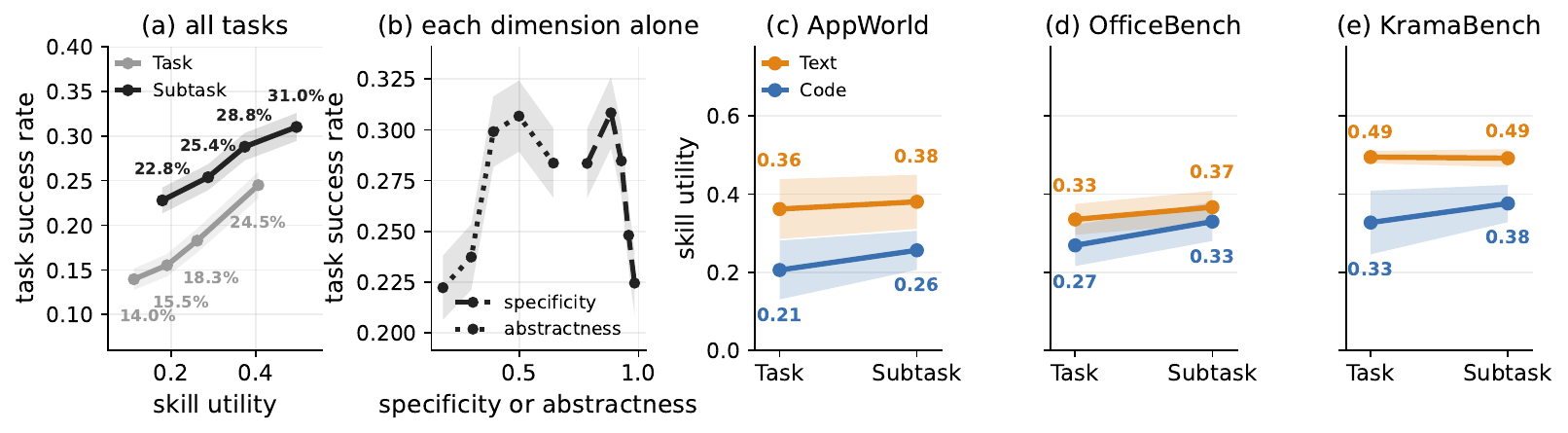}
\caption{Skill utility against the results.
(a): Each task is scored by the average utility of the agent-retrieved skills and ranked into equal-size bins per agent. Success rises from the lowest bin to the highest for both agents.
(b): The subtask-level agent's tasks are scored by one dimension alone and split into equal-size bins, and success rises and then falls on both dimensions.
(c)--(e): The median skill utility of each induced memory, averaged over models, is no lower at the subtask level than the task level, and higher for Text than Code.
Shaded bands are 95\% CIs.}
\label{fig:score-quality}
\end{figure*}

\subsection{A Score for Skill Utility}
\label{sec:why:score}

% Intro: two requirements; notation.
Existing literature hypothesizes a tension between a skill's specificity and its generalizability \citep{yu2025polyskill, fang2026mempexploringagentprocedural}. We suggest that a reusable skill must meet two requirements, specificity and abstractness. Suppose $c_j$ denotes the cosine similarity between a skill $s$ and the $j$th of the $N$ task instructions under the retrieval embedder.

% Specificity: parameter-free probability against the task-pair distribution.
\noindent\textbf{Specificity} measures how close a skill is to the tasks it most resembles. Following \citet{ethayarajh-2019-contextual}, we compare the skill's nearest-task similarity with the similarities between tasks,
\begin{equation*}
\mathrm{specificity}(s) = \Pr\big[\max_j c_j \ge \cos(t_i, t_k)\big],
\end{equation*}
using the probability that the skill is closer to its nearest task than two random tasks $t_i$ and $t_k$ of the benchmark are to each other. Specificity punishes an irrelevant skill far from every task, which scores near zero, and rewards a skill matching at least one task, which scores near one.

% abstractness: effective number of covered tasks, entropy-based generality.
\noindent\textbf{Abstractness} measures how evenly a skill's relevance spreads over many tasks set rather than concentrating on a few tasks. We turn the similarity $c$ into a distribution and compute the perplexity over the task count to represent the ratio of relevant tasks to a skill \citep{JMLR:v9:vandermaaten08a},
\begin{equation*}
\mathrm{abstractness}(s) = \frac{\exp H\big(\mathrm{softmax}(c/\tau)\big)}{N},
\end{equation*}
where $c=(c_1,\dots,c_N)$, $H$ is the entropy, and $\tau=0.1$ is a temperature. Abstractness punishes a skill that is close to a few tasks but far from the rest, and rewards a skill that stays relevant across many tasks, which scores near one.

% Skill utility: tradeoff as motivation, tf-idf analogy, then the product.
\SmallHeading{Skill utility} We show in \Appendix~\ref{app:utility-tradeoff} that specificity and abstractness trade off for both task-level and subtask-level skills. We hypothesize that a useful skill balances the two properties well, so we define skill utility as the product of both properties:
\begin{equation*}
\mathrm{utility}(s) = \mathrm{specificity}(s) \cdot \mathrm{abstractness}(s).
\end{equation*}

\subsection{Skill Utility Predicts Task Success Gains}
\label{sec:why:explain}

% Intro: recap results.
We examine whether skill utility explains the results of \Section~\ref{sec:results}, that subtask skills help mostly and that text skills transfer better than code skills.

% Validation: success rises with the utility of the retrieved skills.
\SmallHeading{Higher utility, more successes} If high-utility skills explain the task success gains, tasks that retrieve higher-utility skills should succeed more. We score each task by the average utility of the skills the agent retrieves when solving the task, rank tasks by the score for task-level and subtask-level agents, and split each ranking into equal-size bins. Success rises monotonically across the bins, from $14.0\%$ to $24.5\%$ for the task-level agent and from $22.8\%$ to $31.0\%$ for the subtask-level agent (\Figure~\ref{fig:score-quality}a).
Over 10 equal-size bins, the Spearman correlation between utility and success is $0.98$ for the task-level agent and $0.95$ for the subtask-level agent, both at $p<0.001$.
The correlation also holds with a second embedder \texttt{all-mpnet-base-v2}~\cite{NEURIPS2020_c3a690be}, different aggregation of specificity and abstractness, and various difficulty levels (\Appendix~\ref{app:utility-robust}).
% Confound check: the utility effect is not task difficulty.
Beyond the correlation between skill utility and task success, we also analyze the causal effect of skill utility. 
We split the same skill library at its median utility and rerun the agent on the same tasks with one half at a time, and the high-utility half yields higher success for both the task-level and the subtask-level library (\Appendix~\ref{app:utility-causal}, \Table~\ref{tab:utility-halves}).

% Ablation: neither dimension alone predicts success; the product does.
\SmallHeading{Neither dimension, by itself, predicts success}
Neither specificity nor abstractness alone predicts the success of the subtask-level agent. In \Figure~\ref{fig:score-quality}b, success rises and then falls as specificity or abstractness increases. The phenomenon follows from the tradeoff in \Figure~\ref{fig:tradeoff}, as a skill high on one dimension gives up the other and loses utility. The agent therefore succeeds most when the two dimensions stay balanced, which the product rewards.

% Winning conditions score higher: subtask > task and text > code.
\SmallHeading{Subtask-level and text skills have higher utility}
% We next compare the skill utility across the two induction levels and the two skill formats. 
% For each condition we take the median utility of induced skills and average the medians over all models on each benchmark.
From \Figure~\ref{fig:score-quality}c to e, subtask-level skills score higher than task-level skills on almost every benchmark and skill format. Besides, text skills score higher than code at both induction levels on every benchmark.

\subsection{Skill Utility Matches Actual Skill Reuse}
\label{sec:why:reuse}

% Intro: from intrinsic score to observed reuse.
Finally, we check whether skill utility reflects the cross-task reuse during the task stream.

\begin{figure}[t]
\centering
\includegraphics[width=\linewidth]{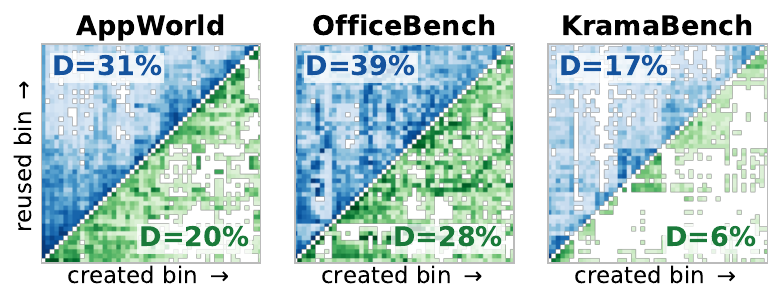}
\caption{Transfer density. The task stream of each benchmark is cut into 50 bins in stream order, and a cell is colored when a skill induced in the earlier bin is retrieved in a later bin, darker when more of the runs do so. Blue upper triangles are the subtask-level agent and green lower triangles are the task-level agent, pooled over models and both skill formats, and the number in each triangle is its transfer density $D$. The subtask-level agent reuses skills more densely than the task-level agent on all three benchmarks.}
\label{fig:transfer}
\end{figure}

% Transfer density: bins + definition + what it reflects.
\SmallHeading{Transfer density} We cut each benchmark's task stream into $b=50$ equal bins in order. Transfer density $D$ is the share of bin pairs that carry at least one actual transfer,
% \begin{equation*}
% D = \frac{1}{\binom{b}{2}} \sum_{i<j} r_{ij},
% \end{equation*}
$D = \sum_{i<j} r_{ij} / \binom{b}{2}$, where $r_{ij}=1$ when a skill induced in bin $i$ is retrieved in a later bin $j$, and $0$ otherwise. 
% We compute the density for each model and skill format and report the average over them.

% Results: denser reuse corroborates the score on the induction-level axis.
\SmallHeading{Results}
The subtask-level agent reuses skills more densely than the task-level agent on all three benchmarks (\Figure~\ref{fig:transfer}), also when divided by the library size (\Appendix~\ref{app:density}). The skill induction level that scores higher utility is thus also the one whose skills are reused more often.

% Conclusion
\section{Conclusion}
\label{sec:conclusion}

% Question, findings, explanation in one short paragraph.
We study when induced skills transfer across tasks, comparing skill induction levels and formats on three benchmarks and eleven models. We show that subtask-level skills can improve the agent while task-level skills harm it, and text skills transfer better than code skills. Skill utility, balancing specificity and abstractness, correlates with task success, ranks the winning conditions higher, and serves as a lightweight diagnostic of a skill memory.

\clearpage

% Mandatory (unnumbered, excluded from the page limit) for *ACL / ARR.
\section*{Limitations}
\label{sec:limitations}

First, our study is restricted to skill transfer in English, text-based agent environments, which span multi-app tool use, office-document workflows, and data-science pipelines, each scored by its official evaluator. These choices give our conclusions a controlled basis.
Consequently, we do not evaluate skill transfer in non-English and multi-modal environments, which may exhibit different skill transfer dynamics.
Second, we conduct experiments on a specific set of open-source and proprietary models (Qwen-3, GPT-OSS, Nemotron, Gemma-3, and Gemini-3.1-Pro). While we include diverse architectures and capabilities (e.g., dense and Mixture-of-Experts models from 4B to 235B parameters), this selection is not exhaustive, and our findings may not strictly generalize to untested models.
% Finally, we score each task by its final environment state, which keeps grading deterministic and comparable. The benchmarks provide no step-level ground truth, so a finer-grained analysis of intermediate decisions would need a judge model, and we leave it to future work.

\section*{Ethical Considerations}
% Risk of malicious skills, why our study is safe, and what is left for future work.
Our work probes when an LLM agent can reliably reuse skills induced from its own past experience. A skill memory that transfers procedures across tasks could also transfer harmful ones, because an adversary who injects malicious skills into the library could steer the agent through the same reuse mechanism. Our experiments carry no such risk, since every skill is induced by the agent itself from sandboxed benchmark tasks without real user data. We regard reliable reuse of trustworthy skills as a prerequisite for the harder challenge of deciding which stored skills to trust, so detecting and rejecting malicious skills is left for future work.
\label{sec:ethics}

\section*{Acknowledgments}
This research was supported by an Amazon Research Award of Spring 2025 on AWS Agentic AI, a Stony Brook Spring 2025 OVPR Seed Grant, and the National Artificial Intelligence Research Resource (NAIRR) Pilot program (award NAIRR250525).
This research used the DeltaAI advanced computing and data resource, which is supported by the National Science Foundation (award OAC 2320345) and the State of Illinois.
% DeltaAI is a joint effort of the University of Illinois Urbana-Champaign and its National Center for Supercomputing Applications.
This work used Amazon Web Services, Google Cloud, and Microsoft Azure through the CloudBank project, which is supported by National Science Foundation grant \#1925001.
We also thank Huajian Zhang for feedback on the writing.

\bibliography{custom}

\begin{thebibliography}{79}
\providecommand{\natexlab}[1]{#1}

\bibitem[{Agarwal et~al.(2025)Agarwal, Ahmad, Ai, Altman, Applebaum, Arbus, Arora, Bai, Baker, Bao et~al.}]{agarwal2025gpt}
Sandhini Agarwal, Lama Ahmad, Jason Ai, Sam Altman, Andy Applebaum, Edwin Arbus, Rahul~K Arora, Yu~Bai, Bowen Baker, Haiming Bao, et~al. 2025.
\newblock \href {https://arxiv.org/abs/2508.10925} {gpt-oss-120b \& gpt-oss-20b model card}.
\newblock \emph{Preprint}, arXiv:2508.10925.

\bibitem[{Ahmed et~al.(2026)Ahmed, Khan, Ahmad, Di, Liu, and Anwar}]{ahmed2026retrievalofthought}
Ammar Ahmed, Azal~Ahmad Khan, Ayaan Ahmad, Sheng Di, Zirui Liu, and Ali Anwar. 2026.
\newblock \href {https://openreview.net/forum?id=Wy7NyScKlD} {Retrieval-of-thought: Efficient reasoning via reusing thoughts}.
\newblock In \emph{The Fourteenth International Conference on Learning Representations}.

\bibitem[{Cai et~al.(2024)Cai, Wang, Ma, Chen, and Zhou}]{cai2024large}
Tianle Cai, Xuezhi Wang, Tengyu Ma, Xinyun Chen, and Denny Zhou. 2024.
\newblock \href {https://openreview.net/forum?id=qV83K9d5WB} {Large language models as tool makers}.
\newblock In \emph{The Twelfth International Conference on Learning Representations}.

\bibitem[{Chandiramani et~al.(2026)Chandiramani, Blakeman, Olaoye, Gupta, Somasamudramath, Khattar, Adesoba, Renduchintala, Asif, Agrawal et~al.}]{nvidia2026nemotron3superopen}
Aakshita Chandiramani, Aaron Blakeman, Abdullahi Olaoye, Abhibha Gupta, Abhilash Somasamudramath, Abhinav Khattar, Adeola Adesoba, Adi Renduchintala, Adil Asif, Aditya Agrawal, et~al. 2026.
\newblock \href {https://arxiv.org/abs/2604.12374} {Nemotron 3 super: Open, efficient mixture-of-experts hybrid mamba-transformer model for agentic reasoning}.
\newblock \emph{Preprint}, arXiv:2604.12374.

\bibitem[{Chen et~al.(2024)Chen, Li, Yang, Yu, Lin, and He}]{chen2024automanual}
Minghao Chen, Yihang Li, Yanting Yang, Shiyu Yu, Binbin Lin, and Xiaofei He. 2024.
\newblock \href {https://doi.org/10.52202/079017-0019} {Automanual: Constructing instruction manuals by llm agents via interactive environmental learning}.
\newblock In \emph{Advances in Neural Information Processing Systems}, volume~37, pages 589--631. Curran Associates, Inc.

\bibitem[{Cuconasu et~al.(2024)Cuconasu, Trappolini, Siciliano, Filice, Campagnano, Maarek, Tonellotto, and Silvestri}]{Cuconasu}
Florin Cuconasu, Giovanni Trappolini, Federico Siciliano, Simone Filice, Cesare Campagnano, Yoelle Maarek, Nicola Tonellotto, and Fabrizio Silvestri. 2024.
\newblock \href {https://doi.org/10.1145/3626772.3657834} {The power of noise: Redefining retrieval for rag systems}.
\newblock In \emph{Proceedings of the 47th International ACM SIGIR Conference on Research and Development in Information Retrieval}, SIGIR '24, page 719–729, New York, NY, USA. Association for Computing Machinery.

\bibitem[{Dua et~al.(2022)Dua, Gupta, Singh, and Gardner}]{dua2022successive}
Dheeru Dua, Shivanshu Gupta, Sameer Singh, and Matt Gardner. 2022.
\newblock \href {https://doi.org/10.18653/v1/2022.emnlp-main.81} {Successive prompting for decomposing complex questions}.
\newblock In \emph{Proceedings of the 2022 Conference on Empirical Methods in Natural Language Processing}, pages 1251--1265, Abu Dhabi, United Arab Emirates. Association for Computational Linguistics.

\bibitem[{Ellis et~al.(2021)Ellis, Wong, Nye, Sabl\'{e}-Meyer, Morales, Hewitt, Cary, Solar-Lezama, and Tenenbaum}]{ellis2021dreamcoder}
Kevin Ellis, Catherine Wong, Maxwell Nye, Mathias Sabl\'{e}-Meyer, Lucas Morales, Luke Hewitt, Luc Cary, Armando Solar-Lezama, and Joshua~B. Tenenbaum. 2021.
\newblock \href {https://doi.org/10.1145/3453483.3454080} {Dreamcoder: bootstrapping inductive program synthesis with wake-sleep library learning}.
\newblock In \emph{Proceedings of the 42nd ACM SIGPLAN International Conference on Programming Language Design and Implementation}, PLDI 2021, page 835–850, New York, NY, USA. Association for Computing Machinery.

\bibitem[{Ethayarajh(2019)}]{ethayarajh-2019-contextual}
Kawin Ethayarajh. 2019.
\newblock \href {https://doi.org/10.18653/v1/D19-1006} {How contextual are contextualized word representations? {C}omparing the geometry of {BERT}, {ELM}o, and {GPT}-2 embeddings}.
\newblock In \emph{Proceedings of the 2019 Conference on Empirical Methods in Natural Language Processing and the 9th International Joint Conference on Natural Language Processing (EMNLP-IJCNLP)}, pages 55--65, Hong Kong, China. Association for Computational Linguistics.

\bibitem[{Fang et~al.(2026)Fang, Liang, Wang, Wu, Qiao, Xie, Huang, Chen, and Zhang}]{fang2026mempexploringagentprocedural}
Runnan Fang, Yuan Liang, Xiaobin Wang, Jialong Wu, Shuofei Qiao, Pengjun Xie, Fei Huang, Huajun Chen, and Ningyu Zhang. 2026.
\newblock \href {https://doi.org/10.18653/v1/2026.findings-acl.866} {Memp: Exploring agent procedural memory}.
\newblock In \emph{Findings of the {A}ssociation for {C}omputational {L}inguistics: {ACL} 2026}, pages 17490--17502, San Diego, California, United States. Association for Computational Linguistics.

\bibitem[{Feng et~al.(2026{\natexlab{a}})Feng, Song, Li, Liu, and Shao}]{feng2026searljointoptimizationpolicy}
Xinshun Feng, Xinhao Song, Lijun Li, Gongshen Liu, and Jing Shao. 2026{\natexlab{a}}.
\newblock \href {https://doi.org/10.18653/v1/2026.acl-long.1125} {{SEARL}: Joint optimization of policy and tool graph memory for self-evolving agents}.
\newblock In \emph{Proceedings of the 64th Annual Meeting of the {A}ssociation for {C}omputational {L}inguistics (Volume 1: Long Papers)}, pages 24518--24535, San Diego, California, United States. Association for Computational Linguistics.

\bibitem[{Feng et~al.(2026{\natexlab{b}})Feng, Chen, Wu, Zhou, and Bosselut}]{feng-etal-2026-tracking}
Yiyang Feng, Zeming Chen, Haotian Wu, Jiawei Zhou, and Antoine Bosselut. 2026{\natexlab{b}}.
\newblock \href {https://doi.org/10.18653/v1/2026.eacl-long.273} {Tracking the limits of knowledge propagation: How {LLM}s fail at multi-step reasoning with conflicting knowledge}.
\newblock In \emph{Proceedings of the 19th Conference of the {E}uropean Chapter of the {A}ssociation for {C}omputational {L}inguistics (Volume 1: Long Papers)}, pages 5813--5847, Rabat, Morocco. Association for Computational Linguistics.

\bibitem[{Feng et~al.(2025)Feng, Wang, Cui, Faltings, Lee, and Zhou}]{feng-etal-2025-unraveling}
Yiyang Feng, Yichen Wang, Shaobo Cui, Boi Faltings, Mina Lee, and Jiawei Zhou. 2025.
\newblock \href {https://doi.org/10.18653/v1/2025.findings-emnlp.627} {Unraveling misinformation propagation in {LLM} reasoning}.
\newblock In \emph{Findings of the Association for Computational Linguistics: EMNLP 2025}, pages 11683--11707, Suzhou, China. Association for Computational Linguistics.

\bibitem[{Fu et~al.(2024)Fu, Kim, Kim, Sohn, Logeswaran, Bae, and Lee}]{fu2024autoguide}
Yao Fu, Dong-Ki Kim, Jaekyeom Kim, Sungryull Sohn, Lajanugen Logeswaran, Kyunghoon Bae, and Honglak Lee. 2024.
\newblock \href {https://openreview.net/forum?id=mRIQz8Zd6O} {Autoguide: Automated generation and selection of context-aware guidelines for large language model agents}.
\newblock In \emph{The Thirty-eighth Annual Conference on Neural Information Processing Systems}.

\bibitem[{Grand et~al.(2024)Grand, Wong, Bowers, Olausson, Liu, Tenenbaum, and Andreas}]{grand2024lilo}
Gabriel Grand, Lionel Wong, Matthew Bowers, Theo~X. Olausson, Muxin Liu, Joshua~B. Tenenbaum, and Jacob Andreas. 2024.
\newblock \href {https://openreview.net/forum?id=TqYbAWKMIe} {{LILO}: Learning interpretable libraries by compressing and documenting code}.
\newblock In \emph{The Twelfth International Conference on Learning Representations}.

\bibitem[{Harris et~al.(2020)Harris, Millman, van~der Walt, Gommers, Virtanen, Cournapeau, Wieser, Taylor, Berg, Smith, Kern, Picus, Hoyer, van Kerkwijk, Brett, Haldane, del R{\'{i}}o, Wiebe, Peterson, G{\'{e}}rard-Marchant, Sheppard, Reddy, Weckesser, Abbasi, Gohlke, and Oliphant}]{harris2020array}
Charles~R. Harris, K.~Jarrod Millman, St{\'{e}}fan~J. van~der Walt, Ralf Gommers, Pauli Virtanen, David Cournapeau, Eric Wieser, Julian Taylor, Sebastian Berg, Nathaniel~J. Smith, Robert Kern, Matti Picus, Stephan Hoyer, Marten~H. van Kerkwijk, Matthew Brett, Allan Haldane, Jaime~Fern{\'{a}}ndez del R{\'{i}}o, Mark Wiebe, Pearu Peterson, and 7 others. 2020.
\newblock \href {https://doi.org/10.1038/s41586-020-2649-2} {Array programming with {NumPy}}.
\newblock \emph{Nature}, 585(7825):357--362.

\bibitem[{Hu et~al.(2025)Hu, Chen, Chen, Mu, Shao, and Luo}]{hu-etal-2025-hiagent}
Mengkang Hu, Tianxing Chen, Qiguang Chen, Yao Mu, Wenqi Shao, and Ping Luo. 2025.
\newblock \href {https://doi.org/10.18653/v1/2025.acl-long.1575} {{H}i{A}gent: Hierarchical working memory management for solving long-horizon agent tasks with large language model}.
\newblock In \emph{Proceedings of the 63rd Annual Meeting of the Association for Computational Linguistics (Volume 1: Long Papers)}, pages 32779--32798, Vienna, Austria. Association for Computational Linguistics.

\bibitem[{Hu et~al.(2026)Hu, Liu, Yue, Zhang, Liu, Zhu, Lin, Guo, Dou, Xi, Jin, Tan, Yin, Liu, Zhang, Sun, Zhu, Sun, Peng, Cheng, Fan, Guo, Yu, Zhou, Hu, Huo, Wang, Niu, Wang, Yin, Hu, Liao, Li, Wang, Zhou, Liu, Cheng, Zhang, Gui, Pan, Zhang, Torr, Dou, Wen, Huang, Jiang, and Yan}]{hu2026memoryageaiagents}
Yuyang Hu, Shichun Liu, Yanwei Yue, Guibin Zhang, Boyang Liu, Fangyi Zhu, Jiahang Lin, Honglin Guo, Shihan Dou, Zhiheng Xi, Senjie Jin, Jiejun Tan, Yanbin Yin, Jiongnan Liu, Zeyu Zhang, Zhongxiang Sun, Yutao Zhu, Hao Sun, Boci Peng, and 28 others. 2026.
\newblock \href {https://arxiv.org/abs/2512.13564} {Memory in the age of ai agents}.
\newblock \emph{Preprint}, arXiv:2512.13564.

\bibitem[{Hunter(2007)}]{Hunter:2007}
J.~D. Hunter. 2007.
\newblock \href {https://doi.org/10.1109/MCSE.2007.55} {Matplotlib: A 2d graphics environment}.
\newblock \emph{Computing in Science \& Engineering}, 9(3):90--95.

\bibitem[{Kaelbling et~al.(1998)Kaelbling, Littman, and Cassandra}]{KAELBLING199899}
Leslie~Pack Kaelbling, Michael~L. Littman, and Anthony~R. Cassandra. 1998.
\newblock \href {https://doi.org/10.1016/S0004-3702(98)00023-X} {Planning and acting in partially observable stochastic domains}.
\newblock \emph{Artificial Intelligence}, 101(1):99--134.

\bibitem[{Kang et~al.(2026)Kang, Cho, and Kim}]{kang2026skillretlargescalebenchmarkskill}
Ryangkyung Kang, Hongcheol Cho, and Youngeun Kim. 2026.
\newblock \href {https://arxiv.org/abs/2605.05726} {Skillret: A large-scale benchmark for skill retrieval in llm agents}.
\newblock \emph{Preprint}, arXiv:2605.05726.

\bibitem[{Kay(2007)}]{1288165.1288167}
Anthony Kay. 2007.
\newblock Tesseract: an open-source optical character recognition engine.
\newblock \emph{Linux J.}, 2007(159):2.

\bibitem[{Khot et~al.(2023)Khot, Trivedi, Finlayson, Fu, Richardson, Clark, and Sabharwal}]{khot2023decomposed}
Tushar Khot, Harsh Trivedi, Matthew Finlayson, Yao Fu, Kyle Richardson, Peter Clark, and Ashish Sabharwal. 2023.
\newblock \href {https://openreview.net/forum?id=_nGgzQjzaRy} {Decomposed prompting: A modular approach for solving complex tasks}.
\newblock In \emph{The Eleventh International Conference on Learning Representations}.

\bibitem[{Kwon et~al.(2023)Kwon, Li, Zhuang, Sheng, Zheng, Yu, Gonzalez, Zhang, and Stoica}]{10.1145/3600006.3613165}
Woosuk Kwon, Zhuohan Li, Siyuan Zhuang, Ying Sheng, Lianmin Zheng, Cody~Hao Yu, Joseph Gonzalez, Hao Zhang, and Ion Stoica. 2023.
\newblock \href {https://doi.org/10.1145/3600006.3613165} {Efficient memory management for large language model serving with pagedattention}.
\newblock In \emph{Proceedings of the 29th Symposium on Operating Systems Principles}, SOSP '23, page 611–626, New York, NY, USA. Association for Computing Machinery.

\bibitem[{Lai et~al.(2026)Lai, Vitagliano, Zhang, Chabra, SUDHIR, Zeng, Zabreyko, Li, Kossmann, Ding, Chen, Markakis, Russo, Wang, Wu, Cafarella, Cao, Madden, and Kraska}]{lai2025kramabench}
Eugenie Lai, Gerardo Vitagliano, Ziyu Zhang, Om~Chabra, SIVAPRASAD SUDHIR, Anna Zeng, Anton~A. Zabreyko, Chenning Li, Ferdi Kossmann, Jialin Ding, Jun Chen, Markos Markakis, Matthew Russo, Weiyang Wang, Ziniu Wu, Mike Cafarella, Lei Cao, Samuel Madden, and Tim Kraska. 2026.
\newblock \href {https://openreview.net/forum?id=fZfUdeCC5X} {{KRAMABENCH}: A benchmark for {AI} systems on data-to-insight pipelines over data lakes}.
\newblock In \emph{The Fourteenth International Conference on Learning Representations}.

\bibitem[{Li et~al.(2026)Li, Zhang, Yu, Duan, Li, Xiang, Liao, Guo, Li, and Suo}]{li2026memposelfmemorypolicyoptimization}
Ruoran Li, Xinghua Zhang, Haiyang Yu, Shitong Duan, Xiang Li, Wenxin Xiang, Chonghua Liao, Xudong Guo, Yongbin Li, and Jinli Suo. 2026.
\newblock \href {https://doi.org/10.18653/v1/2026.findings-acl.1166} {{M}em{PO}: Self-memory policy optimization for long-horizon agents}.
\newblock In \emph{Findings of the {A}ssociation for {C}omputational {L}inguistics: {ACL} 2026}, pages 23286--23301, San Diego, California, United States. Association for Computational Linguistics.

\bibitem[{Liu et~al.(2024)Liu, Lin, Hewitt, Paranjape, Bevilacqua, Petroni, and Liang}]{liu-etal-2024-lost}
Nelson~F. Liu, Kevin Lin, John Hewitt, Ashwin Paranjape, Michele Bevilacqua, Fabio Petroni, and Percy Liang. 2024.
\newblock \href {https://doi.org/10.1162/tacl_a_00638} {Lost in the middle: How language models use long contexts}.
\newblock \emph{Transactions of the Association for Computational Linguistics}, 12:157--173.

\bibitem[{Majumder et~al.(2024)Majumder, Mishra, Jansen, Tafjord, Tandon, Zhang, Callison-Burch, and Clark}]{majumder2024clin}
Bodhisattwa~Prasad Majumder, Bhavana~Dalvi Mishra, Peter Jansen, Oyvind Tafjord, Niket Tandon, Li~Zhang, Chris Callison-Burch, and Peter Clark. 2024.
\newblock \href {https://openreview.net/forum?id=xS6zx1aBI9} {{CLIN}: A continually learning language agent for rapid task adaptation and generalization}.
\newblock In \emph{First Conference on Language Modeling}.

\bibitem[{Nguyen et~al.(2025)Nguyen, Lai, Yoon, Rossi, Zhao, Zhang, Mathur, Lipka, Wang, Bui, Dernoncourt, and Zhou}]{nguyen2025dynasaur}
Dang Nguyen, Viet~Dac Lai, Seunghyun Yoon, Ryan~A. Rossi, Handong Zhao, Ruiyi Zhang, Puneet Mathur, Nedim Lipka, Yu~Wang, Trung Bui, Franck Dernoncourt, and Tianyi Zhou. 2025.
\newblock \href {https://openreview.net/forum?id=lv0cJ2pWVd} {Dynasaur: Large language agents beyond predefined actions}.
\newblock In \emph{Second Conference on Language Modeling}.

\bibitem[{Nottingham et~al.(2024)Nottingham, Majumder, Mishra, Singh, Clark, and Fox}]{nottingham2024skill}
Kolby Nottingham, Bodhisattwa~Prasad Majumder, Bhavana~Dalvi Mishra, Sameer Singh, Peter Clark, and Roy Fox. 2024.
\newblock \href {https://openreview.net/forum?id=9laB7ytoMp} {Skill set optimization: Reinforcing language model behavior via transferable skills}.
\newblock In \emph{Forty-first International Conference on Machine Learning}.

\bibitem[{Ouyang et~al.(2026)Ouyang, Yan, Hsu, Chen, Jiang, Wang, Han, Le, Daruki, Tang, Tirumalashetty, Lee, Rofouei, Lin, Han, Lee, and Pfister}]{ouyang2026reasoningbank}
Siru Ouyang, Jun Yan, I-Hung Hsu, Yanfei Chen, Ke~Jiang, Zifeng Wang, Rujun Han, Long Le, Samira Daruki, Xiangru Tang, Vishy Tirumalashetty, George Lee, Mahsan Rofouei, Hangfei Lin, Jiawei Han, Chen-Yu Lee, and Tomas Pfister. 2026.
\newblock \href {https://openreview.net/forum?id=jL7fwchScm} {Reasoningbank: Scaling agent self-evolving with reasoning memory}.
\newblock In \emph{The Fourteenth International Conference on Learning Representations}.

\bibitem[{Park et~al.(2023)Park, O'Brien, Cai, Morris, Liang, and Bernstein}]{10.1145/3586183.3606763}
Joon~Sung Park, Joseph O'Brien, Carrie~Jun Cai, Meredith~Ringel Morris, Percy Liang, and Michael~S. Bernstein. 2023.
\newblock \href {https://doi.org/10.1145/3586183.3606763} {Generative agents: Interactive simulacra of human behavior}.
\newblock In \emph{Proceedings of the 36th Annual ACM Symposium on User Interface Software and Technology}, UIST '23, New York, NY, USA. Association for Computing Machinery.

\bibitem[{Prasad et~al.(2024)Prasad, Koller, Hartmann, Clark, Sabharwal, Bansal, and Khot}]{prasad-etal-2024-adapt}
Archiki Prasad, Alexander Koller, Mareike Hartmann, Peter Clark, Ashish Sabharwal, Mohit Bansal, and Tushar Khot. 2024.
\newblock \href {https://doi.org/10.18653/v1/2024.findings-naacl.264} {{AD}a{PT}: As-needed decomposition and planning with language models}.
\newblock In \emph{Findings of the Association for Computational Linguistics: NAACL 2024}, pages 4226--4252, Mexico City, Mexico. Association for Computational Linguistics.

\bibitem[{Qian et~al.(2023)Qian, Han, Fung, Qin, Liu, and Ji}]{qian-etal-2023-creator}
Cheng Qian, Chi Han, Yi~Fung, Yujia Qin, Zhiyuan Liu, and Heng Ji. 2023.
\newblock \href {https://doi.org/10.18653/v1/2023.findings-emnlp.462} {{CREATOR}: Tool creation for disentangling abstract and concrete reasoning of large language models}.
\newblock In \emph{Findings of the Association for Computational Linguistics: EMNLP 2023}, pages 6922--6939, Singapore. Association for Computational Linguistics.

\bibitem[{Reimers and Gurevych(2019)}]{reimers-gurevych-2019-sentence}
Nils Reimers and Iryna Gurevych. 2019.
\newblock \href {https://doi.org/10.18653/v1/D19-1410} {Sentence-{BERT}: Sentence embeddings using {S}iamese {BERT}-networks}.
\newblock In \emph{Proceedings of the 2019 Conference on Empirical Methods in Natural Language Processing and the 9th International Joint Conference on Natural Language Processing (EMNLP-IJCNLP)}, pages 3982--3992, Hong Kong, China. Association for Computational Linguistics.

\bibitem[{Sarch et~al.(2024)Sarch, Jang, Tarr, Cohen, Marino, and Fragkiadaki}]{sarch2024vlm}
Gabriel~Herbert Sarch, Lawrence Jang, Michael~J. Tarr, William~W. Cohen, Kenneth Marino, and Katerina Fragkiadaki. 2024.
\newblock \href {https://openreview.net/forum?id=5G7MRfPngt} {{VLM} agents generate their own memories: Distilling experience into embodied programs of thought}.
\newblock In \emph{The Thirty-eighth Annual Conference on Neural Information Processing Systems}.

\bibitem[{Sharma et~al.(2022)Sharma, Torralba, and Andreas}]{sharma-etal-2022-skill}
Pratyusha Sharma, Antonio Torralba, and Jacob Andreas. 2022.
\newblock \href {https://doi.org/10.18653/v1/2022.acl-long.120} {Skill induction and planning with latent language}.
\newblock In \emph{Proceedings of the 60th Annual Meeting of the Association for Computational Linguistics (Volume 1: Long Papers)}, pages 1713--1726, Dublin, Ireland. Association for Computational Linguistics.

\bibitem[{Shen et~al.(2026)Shen, Zhang, Sun, Zeng, and Yue}]{shen2026structurallyalignedsubtasklevelmemory}
Kangning Shen, Jingyuan Zhang, Chenxi Sun, Wencong Zeng, and Yang Yue. 2026.
\newblock \href {https://arxiv.org/abs/2602.21611} {Structurally aligned subtask-level memory for software engineering agents}.
\newblock \emph{Preprint}, arXiv:2602.21611.

\bibitem[{Shi et~al.(2023)Shi, Chen, Misra, Scales, Dohan, Chi, Sch\"{a}rli, and Zhou}]{pmlr-v202-shi23a}
Freda Shi, Xinyun Chen, Kanishka Misra, Nathan Scales, David Dohan, Ed~H. Chi, Nathanael Sch\"{a}rli, and Denny Zhou. 2023.
\newblock \href {https://proceedings.mlr.press/v202/shi23a.html} {Large language models can be easily distracted by irrelevant context}.
\newblock In \emph{Proceedings of the 40th International Conference on Machine Learning}, volume 202 of \emph{Proceedings of Machine Learning Research}, pages 31210--31227. PMLR.

\bibitem[{Shinn et~al.(2023)Shinn, Cassano, Gopinath, Narasimhan, and Yao}]{shinn2023reflexion}
Noah Shinn, Federico Cassano, Ashwin Gopinath, Karthik~R Narasimhan, and Shunyu Yao. 2023.
\newblock \href {https://openreview.net/forum?id=vAElhFcKW6} {Reflexion: language agents with verbal reinforcement learning}.
\newblock In \emph{Thirty-seventh Conference on Neural Information Processing Systems}.

\bibitem[{Song et~al.(2020)Song, Tan, Qin, Lu, and Liu}]{NEURIPS2020_c3a690be}
Kaitao Song, Xu~Tan, Tao Qin, Jianfeng Lu, and Tie-Yan Liu. 2020.
\newblock \href {https://proceedings.neurips.cc/paper_files/paper/2020/file/c3a690be93aa602ee2dc0ccab5b7b67e-Paper.pdf} {Mpnet: Masked and permuted pre-training for language understanding}.
\newblock In \emph{Advances in Neural Information Processing Systems}, volume~33, pages 16857--16867. Curran Associates, Inc.

\bibitem[{Su et~al.(2026)Su, Long, Ai, He, Tang, Wang, Tu, Wang, and Liu}]{su2026skillretrievalaugmentationagentic}
Weihang Su, Jianming Long, Qingyao Ai, Qiaozhi He, Yichen Tang, Changyue Wang, Yiteng Tu, Yingbo Wang, and Yiqun Liu. 2026.
\newblock \href {https://arxiv.org/abs/2604.24594} {Skill retrieval augmentation for agentic ai}.
\newblock \emph{Preprint}, arXiv:2604.24594.

\bibitem[{Sun et~al.(2023)Sun, Zhuang, Kong, Dai, and Zhang}]{sun2023adaplanner}
Haotian Sun, Yuchen Zhuang, Lingkai Kong, Bo~Dai, and Chao Zhang. 2023.
\newblock \href {http://papers.nips.cc/paper\_files/paper/2023/hash/b5c8c1c117618267944b2617add0a766-Abstract-Conference.html} {Adaplanner: Adaptive planning from feedback with language models}.
\newblock In \emph{Advances in Neural Information Processing Systems 36: Annual Conference on Neural Information Processing Systems 2023, NeurIPS 2023, New Orleans, LA, USA, December 10 - 16, 2023}.

\bibitem[{Sun et~al.(2026)Sun, Lu, Ling, Liu, Yao, Yang, and Chen}]{sun2026scaling}
Weiwei Sun, Miao Lu, Zhan Ling, Kang Liu, Xuesong Yao, Yiming Yang, and Jiecao Chen. 2026.
\newblock \href {https://openreview.net/forum?id=lNRgWoGfYg} {Scaling long-horizon agent via context folding}.
\newblock In \emph{Forty-third International Conference on Machine Learning}.

\bibitem[{Tan et~al.(2025)Tan, Zhang, Ma, Chen, Dai, and Dong}]{tan-etal-2025-membench}
Haoran Tan, Zeyu Zhang, Chen Ma, Xu~Chen, Quanyu Dai, and Zhenhua Dong. 2025.
\newblock \href {https://doi.org/10.18653/v1/2025.findings-acl.989} {{M}em{B}ench: Towards more comprehensive evaluation on the memory of {LLM}-based agents}.
\newblock In \emph{Findings of the Association for Computational Linguistics: ACL 2025}, pages 19336--19352, Vienna, Austria. Association for Computational Linguistics.

\bibitem[{Team et~al.(2025)Team, Kamath, Ferret, Pathak, Vieillard, Merhej, Perrin, Matejovicova, Ramé, Rivière et~al.}]{gemmateam2025gemma3technicalreport}
Gemma Team, Aishwarya Kamath, Johan Ferret, Shreya Pathak, Nino Vieillard, Ramona Merhej, Sarah Perrin, Tatiana Matejovicova, Alexandre Ramé, Morgane Rivière, et~al. 2025.
\newblock \href {https://arxiv.org/abs/2503.19786} {Gemma 3 technical report}.
\newblock \emph{Preprint}, arXiv:2503.19786.

\bibitem[{Trivedi et~al.(2024)Trivedi, Khot, Hartmann, Manku, Dong, Li, Gupta, Sabharwal, and Balasubramanian}]{trivedi-etal-2024-appworld}
Harsh Trivedi, Tushar Khot, Mareike Hartmann, Ruskin Manku, Vinty Dong, Edward Li, Shashank Gupta, Ashish Sabharwal, and Niranjan Balasubramanian. 2024.
\newblock \href {https://doi.org/10.18653/v1/2024.acl-long.850} {{A}pp{W}orld: A controllable world of apps and people for benchmarking interactive coding agents}.
\newblock In \emph{Proceedings of the 62nd Annual Meeting of the Association for Computational Linguistics (Volume 1: Long Papers)}, pages 16022--16076, Bangkok, Thailand. Association for Computational Linguistics.

\bibitem[{van~der Maaten and Hinton(2008)}]{JMLR:v9:vandermaaten08a}
Laurens van~der Maaten and Geoffrey Hinton. 2008.
\newblock \href {http://jmlr.org/papers/v9/vandermaaten08a.html} {Visualizing data using t-sne}.
\newblock \emph{Journal of Machine Learning Research}, 9(86):2579--2605.

\bibitem[{Virtanen et~al.(2020)Virtanen, Gommers, Oliphant, Haberland, Reddy, Cournapeau, Burovski, Peterson, Weckesser, Bright, {van der Walt}, Brett, Wilson, Millman, Mayorov, Nelson, Jones, Kern, Larson, Carey, Polat, Feng, Moore, {VanderPlas}, Laxalde, Perktold, Cimrman, Henriksen, Quintero, Harris, Archibald, Ribeiro, Pedregosa, {van Mulbregt}, and {SciPy 1.0 Contributors}}]{2020SciPy-NMeth}
Pauli Virtanen, Ralf Gommers, Travis~E. Oliphant, Matt Haberland, Tyler Reddy, David Cournapeau, Evgeni Burovski, Pearu Peterson, Warren Weckesser, Jonathan Bright, St{\'e}fan~J. {van der Walt}, Matthew Brett, Joshua Wilson, K.~Jarrod Millman, Nikolay Mayorov, Andrew R.~J. Nelson, Eric Jones, Robert Kern, Eric Larson, and 16 others. 2020.
\newblock \href {https://doi.org/10.1038/s41592-019-0686-2} {{{SciPy} 1.0: Fundamental Algorithms for Scientific Computing in Python}}.
\newblock \emph{Nature Methods}, 17:261--272.

\bibitem[{Wang et~al.(2024{\natexlab{a}})Wang, Xie, Jiang, Mandlekar, Xiao, Zhu, Fan, and Anandkumar}]{wang2024voyager}
Guanzhi Wang, Yuqi Xie, Yunfan Jiang, Ajay Mandlekar, Chaowei Xiao, Yuke Zhu, Linxi Fan, and Anima Anandkumar. 2024{\natexlab{a}}.
\newblock \href {https://openreview.net/forum?id=ehfRiF0R3a} {Voyager: An open-ended embodied agent with large language models}.
\newblock \emph{Transactions on Machine Learning Research}.

\bibitem[{Wang et~al.(2026)Wang, Yan, Wang, Tian, Mishra, Xu, Gandhi, Xu, and Cheong}]{wang2026reinforcementlearningselfimprovingagent}
Jiongxiao Wang, Qiaojing Yan, Yawei Wang, Yijun Tian, Soumya~Smruti Mishra, Zhichao Xu, Megha Gandhi, Panpan Xu, and Lin~Lee Cheong. 2026.
\newblock \href {https://arxiv.org/abs/2512.17102} {Reinforcement learning for self-improving agent with skill library}.
\newblock \emph{Preprint}, arXiv:2512.17102.

\bibitem[{Wang et~al.(2025{\natexlab{a}})Wang, Xu, Wang, Zhang, Yan, Zhang, Huang, and Ji}]{wang2025mobileagente}
Zhenhailong Wang, Haiyang Xu, Junyang Wang, Xi~Zhang, Ming Yan, Ji~Zhang, Fei Huang, and Heng Ji. 2025{\natexlab{a}}.
\newblock \href {https://openreview.net/forum?id=GRmrGws6Lf} {Mobile-agent-e: Self-evolving mobile assistant for complex tasks}.
\newblock In \emph{Workshop on Scaling Environments for Agents}.

\bibitem[{Wang et~al.(2024{\natexlab{b}})Wang, Neubig, and Fried}]{wang2024trove}
Zhiruo Wang, Graham Neubig, and Daniel Fried. 2024{\natexlab{b}}.
\newblock \href {https://openreview.net/forum?id=DCNCwaMJjI} {Tro{VE}: Inducing verifiable and efficient toolboxes for solving programmatic tasks}.
\newblock In \emph{Forty-first International Conference on Machine Learning}.

\bibitem[{Wang et~al.(2024{\natexlab{c}})Wang, Cui, Zhong, Zhang, Yin, Lin, and Shang}]{wang2024officebench}
Zilong Wang, Yuedong Cui, Li~Zhong, Zimin Zhang, Da~Yin, Bill~Yuchen Lin, and Jingbo Shang. 2024{\natexlab{c}}.
\newblock \href {https://arxiv.org/abs/2407.19056} {Officebench: Benchmarking language agents across multiple applications for office automation}.
\newblock \emph{Preprint}, arXiv:2407.19056.

\bibitem[{Wang et~al.(2025{\natexlab{b}})Wang, Gandhi, Neubig, and Fried}]{wang2025inducing}
Zora~Zhiruo Wang, Apurva Gandhi, Graham Neubig, and Daniel Fried. 2025{\natexlab{b}}.
\newblock \href {https://openreview.net/forum?id=lsAY6fWsog} {Inducing programmatic skills for agentic tasks}.
\newblock In \emph{Second Conference on Language Modeling}.

\bibitem[{Wang et~al.(2025{\natexlab{c}})Wang, Mao, Fried, and Neubig}]{wang2025agent}
Zora~Zhiruo Wang, Jiayuan Mao, Daniel Fried, and Graham Neubig. 2025{\natexlab{c}}.
\newblock \href {https://openreview.net/forum?id=NTAhi2JEEE} {Agent workflow memory}.
\newblock In \emph{Forty-second International Conference on Machine Learning}.

\bibitem[{{W}es {M}c{K}inney(2010)}]{mckinney-proc-scipy-2010}
{W}es {M}c{K}inney. 2010.
\newblock \href {https://doi.org/10.25080/Majora-92bf1922-00a} {{D}ata {S}tructures for {S}tatistical {C}omputing in {P}ython}.
\newblock In \emph{{P}roceedings of the 9th {P}ython in {S}cience {C}onference}, pages 56 -- 61.

\bibitem[{Xia et~al.(2026)Xia, Chen, Wang, Liu, Zeng, Wang, Han, Zhou, Zhao, Chen, Zheng, Xie, and Yao}]{xia2026skillrl}
Peng Xia, Jianwen Chen, Hanyang Wang, Jiaqi Liu, Kaide Zeng, Yu~Wang, Siwei Han, Yiyang Zhou, Xujiang Zhao, Haifeng Chen, Zeyu Zheng, Cihang Xie, and Huaxiu Yao. 2026.
\newblock \href {https://openreview.net/forum?id=FYc2IygegR} {Skill{RL}: Evolving agents via recursive skill-augmented reinforcement learning}.
\newblock In \emph{ICLR 2026 Workshop on Lifelong Agents: Learning, Aligning, Evolving}.

\bibitem[{Xiong et~al.(2026)Xiong, Lin, Xie, He, Liu, Tang, Lakkaraju, and Xiang}]{xiong-etal-2026-memory}
Zidi Xiong, Yuping Lin, Wenya Xie, Pengfei He, Zirui Liu, Jiliang Tang, Himabindu Lakkaraju, and Zhen Xiang. 2026.
\newblock \href {https://doi.org/10.18653/v1/2026.acl-long.27} {How memory management impacts {LLM} agents: An empirical study of experience-following behavior}.
\newblock In \emph{Proceedings of the 64th Annual Meeting of the {A}ssociation for {C}omputational {L}inguistics (Volume 1: Long Papers)}, pages 623--645, San Diego, California, United States. Association for Computational Linguistics.

\bibitem[{Yan et~al.(2026)Yan, Yang, Huang, Nie, Ding, Li, Ma, Bi, Kersting, Pan, Schuetze, Tresp, and Ma}]{yan-etal-2026-memory}
Sikuan Yan, Xiufeng Yang, Zuchao Huang, Ercong Nie, Zifeng Ding, Zonggen Li, Xiaowen Ma, Jinhe Bi, Kristian Kersting, Jeff~Z. Pan, Hinrich Schuetze, Volker Tresp, and Yunpu Ma. 2026.
\newblock \href {https://doi.org/10.18653/v1/2026.acl-long.583} {Memory-r1: Enhancing large language model agents to manage and utilize memories via reinforcement learning}.
\newblock In \emph{Proceedings of the 64th Annual Meeting of the {A}ssociation for {C}omputational {L}inguistics (Volume 1: Long Papers)}, pages 12805--12825, San Diego, California, United States. Association for Computational Linguistics.

\bibitem[{Yang et~al.(2025)Yang, Li, Yang, Zhang, Hui, Zheng, Yu, Gao, Huang, Lv et~al.}]{yang2025qwen3}
An~Yang, Anfeng Li, Baosong Yang, Beichen Zhang, Binyuan Hui, Bo~Zheng, Bowen Yu, Chang Gao, Chengen Huang, Chenxu Lv, et~al. 2025.
\newblock \href {https://arxiv.org/abs/2505.09388} {Qwen3 technical report}.
\newblock \emph{Preprint}, arXiv:2505.09388.

\bibitem[{Yang et~al.(2026)Yang, Yang, Wen, Fu, Mei, Wu, Cai, Shen, Deng, Xu, Shi, Qiao, and Li}]{yang-etal-2026-towards}
Cheng Yang, Xuemeng Yang, Licheng Wen, Daocheng Fu, Jianbiao Mei, Rong Wu, Pinlong Cai, Yufan Shen, Nianchen Deng, Jia Xu, Botian Shi, Yu~Qiao, and Haifeng Li. 2026.
\newblock \href {https://doi.org/10.18653/v1/2026.findings-acl.1522} {Towards self-evolving agents: Enabling autonomy through interactive experience refinement}.
\newblock In \emph{Findings of the {A}ssociation for {C}omputational {L}inguistics: {ACL} 2026}, pages 30424--30451, San Diego, California, United States. Association for Computational Linguistics.

\bibitem[{Yang et~al.(2024)Yang, Yu, Zhang, Cao, Xu, Zhang, Gonzalez, and CUI}]{yang2024buffer}
Ling Yang, Zhaochen Yu, Tianjun Zhang, Shiyi Cao, Minkai Xu, Wentao Zhang, Joseph~E. Gonzalez, and Bin CUI. 2024.
\newblock \href {https://openreview.net/forum?id=ANO1i9JPtb} {Buffer of thoughts: Thought-augmented reasoning with large language models}.
\newblock In \emph{The Thirty-eighth Annual Conference on Neural Information Processing Systems}.

\bibitem[{Yao et~al.(2023)Yao, Zhao, Yu, Du, Shafran, Narasimhan, and Cao}]{yao2023react}
Shunyu Yao, Jeffrey Zhao, Dian Yu, Nan Du, Izhak Shafran, Karthik~R Narasimhan, and Yuan Cao. 2023.
\newblock \href {https://openreview.net/forum?id=WE_vluYUL-X} {React: Synergizing reasoning and acting in language models}.
\newblock In \emph{The Eleventh International Conference on Learning Representations}.

\bibitem[{Ye et~al.(2026)Ye, Zhang, Li, Yin, Tao, Zhao, Su, Zhang, Qiao, Wang, Xie, Huang, Zhou, Chen, and Jiang}]{ye2026agentfold}
Rui Ye, Zhongwang Zhang, Kuan Li, Huifeng Yin, Zhengwei Tao, Yida Zhao, Liangcai Su, Liwen Zhang, Zile Qiao, Xinyu Wang, Pengjun Xie, Fei Huang, Jingren Zhou, Siheng Chen, and Yong Jiang. 2026.
\newblock \href {https://openreview.net/forum?id=IuZoTgsUws} {Agentfold: Long-horizon web agents with proactive context folding}.
\newblock In \emph{The Fourteenth International Conference on Learning Representations}.

\bibitem[{Yoran et~al.(2024)Yoran, Wolfson, Ram, and Berant}]{yoran2024making}
Ori Yoran, Tomer Wolfson, Ori Ram, and Jonathan Berant. 2024.
\newblock \href {https://openreview.net/forum?id=ZS4m74kZpH} {Making retrieval-augmented language models robust to irrelevant context}.
\newblock In \emph{The Twelfth International Conference on Learning Representations}.

\bibitem[{Yu et~al.(2025)Yu, Li, Shi, and Qi}]{yu2025polyskill}
Simon Yu, Gang Li, Weiyan Shi, and Peng Qi. 2025.
\newblock \href {https://arxiv.org/abs/2510.15863} {Polyskill: Learning generalizable skills through polymorphic abstraction}.
\newblock \emph{Preprint}, arXiv:2510.15863.

\bibitem[{Yuan et~al.(2024)Yuan, Chen, Wang, Fung, Peng, and Ji}]{yuan2024craft}
Lifan Yuan, Yangyi Chen, Xingyao Wang, Yi~Fung, Hao Peng, and Heng Ji. 2024.
\newblock \href {https://openreview.net/forum?id=G0vdDSt9XM} {{CRAFT}: Customizing {LLM}s by creating and retrieving from specialized toolsets}.
\newblock In \emph{The Twelfth International Conference on Learning Representations}.

\bibitem[{Zhang et~al.(2026)Zhang, Wang, Zhou, Liao, Feng, Li, Zheng, Zhang, Wen, Li, Xiong, Qi, Tang, and Wen}]{zhang2026memrlselfevolvingagentsruntime}
Shengtao Zhang, Jiaqian Wang, Ruiwen Zhou, Junwei Liao, Yuchen Feng, Zhuo Li, Yujie Zheng, Weinan Zhang, Ying Wen, Zhiyu Li, Feiyu Xiong, Yutao Qi, Bo~Tang, and Muning Wen. 2026.
\newblock \href {https://arxiv.org/abs/2601.03192} {Memrl: Self-evolving agents via runtime reinforcement learning on episodic memory}.
\newblock \emph{Preprint}, arXiv:2601.03192.

\bibitem[{Zhao et~al.(2024)Zhao, Huang, Xu, Lin, Liu, and Huang}]{Zhao_Huang_Xu_Lin_Liu_Huang_2024}
Andrew Zhao, Daniel Huang, Quentin Xu, Matthieu Lin, Yong-Jin Liu, and Gao Huang. 2024.
\newblock \href {https://doi.org/10.1609/aaai.v38i17.29936} {Expel: Llm agents are experiential learners}.
\newblock \emph{Proceedings of the AAAI Conference on Artificial Intelligence}, 38(17):19632--19642.

\bibitem[{Zheng et~al.(2025)Zheng, Fatemi, Jin, Wang, Gandhi, Song, Gu, Srinivasa, Liu, Neubig, and Su}]{zheng2025skillweaverwebagentsselfimprove}
Boyuan Zheng, Michael~Y. Fatemi, Xiaolong Jin, Zora~Zhiruo Wang, Apurva Gandhi, Yueqi Song, Yu~Gu, Jayanth Srinivasa, Gaowen Liu, Graham Neubig, and Yu~Su. 2025.
\newblock \href {https://arxiv.org/abs/2504.07079} {Skillweaver: Web agents can self-improve by discovering and honing skills}.
\newblock \emph{Preprint}, arXiv:2504.07079.

\bibitem[{Zheng et~al.(2024)Zheng, Wang, Wang, and An}]{zheng2024synapse}
Longtao Zheng, Rundong Wang, Xinrun Wang, and Bo~An. 2024.
\newblock \href {https://openreview.net/forum?id=Pc8AU1aF5e} {Synapse: Trajectory-as-exemplar prompting with memory for computer control}.
\newblock In \emph{The Twelfth International Conference on Learning Representations}.

\bibitem[{Zhong et~al.(2026)Zhong, Lu, Ning, Wan, Feng, Ao, Ribeiro, Dreyer, Ammirati, and Xiong}]{zhong2026skilllearnbenchbenchmarkingcontinuallearning}
Shanshan Zhong, Yi~Lu, Jingjie Ning, Yibing Wan, Lihan Feng, Yuyi Ao, Leonardo F.~R. Ribeiro, Markus Dreyer, Sean Ammirati, and Chenyan Xiong. 2026.
\newblock \href {https://arxiv.org/abs/2604.20087} {Skilllearnbench: Benchmarking continual learning methods for agent skill generation on real-world tasks}.
\newblock \emph{Preprint}, arXiv:2604.20087.

\bibitem[{Zhong et~al.(2023)Zhong, Wu, Manning, Potts, and Chen}]{zhong-etal-2023-mquake}
Zexuan Zhong, Zhengxuan Wu, Christopher Manning, Christopher Potts, and Danqi Chen. 2023.
\newblock \href {https://doi.org/10.18653/v1/2023.emnlp-main.971} {{MQ}u{AKE}: Assessing knowledge editing in language models via multi-hop questions}.
\newblock In \emph{Proceedings of the 2023 Conference on Empirical Methods in Natural Language Processing}, pages 15686--15702, Singapore. Association for Computational Linguistics.

\bibitem[{Zhou et~al.(2023)Zhou, Sch{\"a}rli, Hou, Wei, Scales, Wang, Schuurmans, Cui, Bousquet, Le, and Chi}]{zhou2023leasttomost}
Denny Zhou, Nathanael Sch{\"a}rli, Le~Hou, Jason Wei, Nathan Scales, Xuezhi Wang, Dale Schuurmans, Claire Cui, Olivier Bousquet, Quoc~V Le, and Ed~H. Chi. 2023.
\newblock \href {https://openreview.net/forum?id=WZH7099tgfM} {Least-to-most prompting enables complex reasoning in large language models}.
\newblock In \emph{The Eleventh International Conference on Learning Representations}.

\bibitem[{Zhou et~al.(2025{\natexlab{a}})Zhou, Chen, Guo, Yan, Lee, Wang, Lee, Zhang, Shao, Yang, and Wang}]{zhou2025mementofinetuningllmagents}
Huichi Zhou, Yihang Chen, Siyuan Guo, Xue Yan, Kin~Hei Lee, Zihan Wang, Ka~Yiu Lee, Guchun Zhang, Kun Shao, Linyi Yang, and Jun Wang. 2025{\natexlab{a}}.
\newblock \href {https://arxiv.org/abs/2508.16153} {Memento: Fine-tuning llm agents without fine-tuning llms}.
\newblock \emph{Preprint}, arXiv:2508.16153.

\bibitem[{Zhou et~al.(2025{\natexlab{b}})Zhou, Yang, Lin, Bai, Zhou, Wang, Levine, and Li}]{zhou2025proposeragentevaluator}
Yifei Zhou, Qianlan Yang, Kaixiang Lin, Min Bai, Xiong Zhou, Yu-Xiong Wang, Sergey Levine, and Li~Erran Li. 2025{\natexlab{b}}.
\newblock \href {https://openreview.net/forum?id=uCkwqAG0uS} {Proposer-agent-evaluator ({PAE}): Autonomous skill discovery for foundation model internet agents}.
\newblock In \emph{Forty-second International Conference on Machine Learning}.

\bibitem[{Zhou et~al.(2026)Zhou, Qu, Wu, Kim, Prakash, Rus, Low, and Liang}]{zhou2026mem}
Zijian Zhou, Ao~Qu, Zhaoxuan Wu, Sunghwan Kim, Alok Prakash, Daniela Rus, Bryan Kian~Hsiang Low, and Paul~Pu Liang. 2026.
\newblock \href {https://openreview.net/forum?id=XY8AaxDSLb} {{MEM}1: Learning to synergize memory and reasoning for efficient long-horizon agents}.
\newblock In \emph{The Fourteenth International Conference on Learning Representations}.

\bibitem[{Zhu et~al.(2023)Zhu, Chen, Tian, Tao, Su, Yang, Huang, Li, Lu, Wang, Qiao, Zhang, and Dai}]{zhu2023ghostminecraftgenerallycapable}
Xizhou Zhu, Yuntao Chen, Hao Tian, Chenxin Tao, Weijie Su, Chenyu Yang, Gao Huang, Bin Li, Lewei Lu, Xiaogang Wang, Yu~Qiao, Zhaoxiang Zhang, and Jifeng Dai. 2023.
\newblock \href {https://arxiv.org/abs/2305.17144} {Ghost in the minecraft: Generally capable agents for open-world environments via large language models with text-based knowledge and memory}.
\newblock \emph{Preprint}, arXiv:2305.17144.

\end{thebibliography}

\clearpage

\appendix

\section*{Table of Contents}
\startcontents[appendix]
\printcontents[appendix]{}{1}{\setcounter{tocdepth}{2}}

\section{Prompts}
\label{app:prompts}

% What this appendix lists and how to read the boxes.
We list the deployed prompts of the two agents and the two skill formats on the three benchmarks. Blue boxes hold system messages and yellow boxes hold user messages. Curly braces mark values filled at run time, and a gray hook marks a soft line wrap.
% The parity guarantee that matters for the comparison.
The two induction levels share the skill induction prompt, and task-level agent is a special case of subtask-level agent because it has the whole task as the only one subgoal, so an induction level differs only in the trajectory span the induction step reads.

\subsection{Subtask-Level Agent}
\label{app:prompt:subtask}

% The executor and the shared execution core.
The subtask-level agent runs a planner, an executor, and a summarizer, and the executor is the role that acts on the environment. The executor's system prompt is the execution core of \Figures~\ref{fig:prompt-core-aw}, \ref{fig:prompt-core-ob}, and \ref{fig:prompt-core-kb}, followed by its role block in \Figures~\ref{fig:prompt-subtask-aw}, \ref{fig:prompt-subtask-ob}, and \ref{fig:prompt-subtask-kb}.

% The executor's user message and retrieval framing.
The executor's user message (\Figure~\ref{fig:prompt-subtask-user}) carries the current subgoal, the task for context, the progress summary, and on OfficeBench the live action menu. The skills retrieved for a subgoal enter this message wrapped in the format framing of \Figure~\ref{fig:prompt-subtask-frame}, so the executor treats remembered procedures as hints for the current subgoal only.

% Planner and summarizer around the executor.
The planner and the summarizer reuse the same execution core and append their own role blocks. The planner is rebuilt fresh at every call, so it sees only the task and the current progress summary. The summarizer reads the executor conversation and replies with one progress summary tool call. Their user messages are also in \Figure~\ref{fig:prompt-subtask-user}.

\subsection{Task-Level Agent}
\label{app:prompt:react}

% The special case whose single subgoal is the whole task.
The task-level agent is the executor run as a special case whose single subgoal is the whole task. The planner and summarizer blocks drop away, and it has the same system and user prompts as the executor of the subtask agent. The only difference is that the single subgoal is the whole task instead of one produced by the planner.
With skill memory on, the retrieved skills enter as one further user message under the framing of \Figure~\ref{fig:prompt-subtask-frame}. A retrieved code skill is also loaded into the execution namespace so the agent can call the function by name.

\subsection{Text-Skill Induction}
\label{app:prompt:text}

% When the prompt fires and what it asks for.
After a task at the task level, or after each subgoal at the subtask level, the induction prompt of \Figures~\ref{fig:prompt-text-aw}, \ref{fig:prompt-text-ob}, or \ref{fig:prompt-text-kb} is appended to the finished span as a user message. The model must answer with one skill induction tool call whose arguments hold the description and the note body.

\subsection{Code-Skill Induction}
\label{app:prompt:code}

% Same protocol with a function payload.
Code-skill induction follows the same protocol with the prompts of \Figures~\ref{fig:prompt-code-aw}, \ref{fig:prompt-code-ob}, and \ref{fig:prompt-code-kb}. The tool call returns a functions list whose single entry holds a name, a description, and a Python implementation, and an empty list is reserved for spans with no real work.

\section{Skill Memory}

\subsection{Examples of Induced Skills}
\label{app:skill-examples}

% Real induced skills from one task, all four conditions, fresh at induction time.
We show real skills induced by Qwen3-235B-A22B from one AppWorld task, which asks the agent to order all weightlifting benches in the Amazon cart. Each skill appears as induced right after its source trajectory, before any later merge. Task-level induction yields one skill for the whole task (\Figures~\ref{fig:skill-example-task-text} and \ref{fig:skill-example-task-code}), and subtask-level induction yields one skill per subtask, three in this run (\Figures~\ref{fig:skill-example-subtask-text} and \ref{fig:skill-example-subtask-code}). The scope contrast is easiest to see in the code format, as the task-level function keeps the source task's product keyword as its default argument while the subtask-level functions stay generic over login, cart filtering, and order placement.

\subsection{Skill Memory Mechanics}
\label{app:memory}
Both memories embed skill descriptions and queries with \texttt{all-MiniLM-L6-v2} and rank candidates by cosine similarity. Retrieval returns the top 5 text or code skills above 0.30. A selected code skill also pulls in the stored functions it calls. A new text skill merges into a stored entry whose description similarity exceeds 0.85, replacing the procedure line and appending only unseen bullets, and otherwise becomes a new entry. A new code skill overwrites the entry whose function name it reuses, is dropped as a near duplicate at description similarity 0.85, and otherwise becomes a new entry. Execution feedback prunes code skills further. A stored function that fails to load into the execution namespace is removed.
% and the subtask-level agent also removes a retrieved function whose call raises an error. 
Each removal cascades to the functions whose source calls the removed one.

\subsection{Validation of Retrieval Quality}
\label{app:selfretr}

We inspect the retriever to validate that the skill effect does not stem from retriever quality. We hypothesize that a skill induced from a task should be retrievable from that same task. Specifically, we compute the cosine similarity between each skill and the task or subtask from which it was induced. A skill counts as \textit{self-retrievable} if this similarity exceeds the minimum similarity between that task or subtask and its originally retrieved skills. In other words, the skill would rank inside the original retrieval set of its source. All conditions use the deployed embedder and the same retrieval thresholds, so both formats and both induction levels face the same bar.

\begin{table}[t]
\centering
\small
\begin{tabular}{lcc}
\toprule
Induction level & +Text & +Code \\
\midrule
Task-level    & 75.6\ci{72.8}{78.3} & 88.1\ci{86.2}{89.9} \\
Subtask-level & 79.6\ci{78.1}{80.9} & 86.1\ci{85.0}{87.2} \\
\bottomrule
\end{tabular}
\caption{Self-retrieval rates on Qwen3-235B-A22B, Nemotron-Super-120B, and Qwen3-32B, pooled over the three benchmarks, with 95\% task-bootstrap confidence intervals. A skill counts as \textit{self-retrievable} if it ranks among the original skills retrieved for its source task or subtask. All conditions use the same retrieval configuration.}
\label{tab:selfretr}
\end{table}

% Result 1: induction level makes no difference, so retrieval cannot explain the granularity gap.
\Table~\ref{tab:selfretr} reports self-retrieval rates on Qwen3-235B-A22B, Nemotron-Super-120B, and Qwen3-32B, pooled over the three benchmarks. First, the two induction levels are statistically indistinguishable, so retrieval quality cannot explain the level effect. Second, Code skills self-retrieve better than Text skills, yet \Section~\ref{sec:results:fmt} shows that Text skills help more than Code skills, so retrieval quality cannot explain the format effect either. We hypothesize that the low self-retrieval rate of Text skills arises because Text descriptions are the most abstract. On KramaBench, nearly half of the task-level Text skills are not self-retrievable because their descriptions keep the reusable wrangling pattern but drop the topical words of the question. In \Section~\ref{sec:why:explain}, however, we show that this same abstraction leads to better cross-task skill transfer.

% Secondary check: an LLM judge finds the induced skills almost always relevant to their source task in every condition.
As an additional post-hoc check, we sample 50 induced skills per condition from the same population and ask Gemini-3.1-Pro whether each skill is relevant to its source task. The judge marks $100$, $100$, $98$, and $100$ percent of the skills as relevant for Task+Text, Task+Code, Subtask+Text, and Subtask+Code, respectively. Induced skills are thus almost always relevant to their source task in every condition, so differences in skill relevance cannot explain the skill effect.

\section{Experimental Details}
\label{app:eval}

\subsection{Model Configuration}
\label{app:models}

% Serving and precision.
We serve Qwen3-235B-A22B, GPT-OSS-120B, and Nemotron-Super-120B on Amazon Bedrock and Gemini-3.1-Pro on Google Cloud. We serve each dense model with vLLM~\citep{10.1145/3600006.3613165} from its released bf16 weights without quantization, on one 96\,GB NVIDIA GH200 with a context window of 40,960 tokens.
% Reasoning.
Thinking is off for the dense Qwen3 models. GPT-OSS-120B and Nemotron-Super-120B run with low reasoning effort, the lowest setting of GPT-OSS-120B, and Gemini-3.1-Pro runs with a low thinking level. Qwen3-235B-A22B is the instruct release, which has no reasoning setting.
% Sampling.
All models keep the provider's default temperature and top-p and truncate generated tokens at 8192 per call. The local models also use a repetition penalty of 1.1, which keeps the small models from repeating the same output.

\subsection{Performance Scoring}
\label{app:eval:perf}

% Shared definition.
Each benchmark's official evaluator maps a finished task to a score in $[0,1]$, and the reported task success of a condition is the mean score over its tasks.
% AppWorld.
(\textit{i}) AppWorld scores a task with its official state-based unit tests, which check the final environment state left by the agent's API calls. The score is 1 when every test passes and 0 otherwise.
% OfficeBench.
(\textit{ii}) OfficeBench attaches official checker functions to each task, which inspect the files and application state the agent leaves in the sandbox. The score is 1 when every checker passes and 0 otherwise.
% KramaBench.
(\textit{iii}) KramaBench grades the submitted answer against the gold answer with the official deterministic scorer of its answer type, and a task that never submits scores 0. Exact numeric and string answers score 0 or 1 under a small numeric tolerance. List answers score the F1 overlap with the gold list, and approximate numeric answers score $1/(1+e)$ for a relative absolute error $e$. The per-task score is therefore continuous, and the few tasks graded by an LLM judge are excluded from our task set.

\begin{table}[t]
\centering
\small
\setlength{\tabcolsep}{4pt}
\begin{tabular}{lcccc}
\toprule
 & \multicolumn{2}{c}{Latency (s)} & \multicolumn{2}{c}{Tokens (k)} \\
\cmidrule(lr){2-3}\cmidrule(lr){4-5}
Model & Task & Subtask & Task & Subtask \\
\midrule
\multicolumn{5}{l}{\textit{Amazon Bedrock}} \\
Qwen3-235B-A22B     & 78  & 120 & 152 & 284 \\
GPT-OSS-120B        & 67  & 88  & 89  & 193 \\
Nemotron-Super-120B & 509 & 613 & 391 & 484 \\
\multicolumn{5}{l}{\textit{Google Cloud}} \\
Gemini-3.1-Pro      & 49  & 100 & 111 & 177 \\
\multicolumn{5}{l}{\textit{vLLM on one GH200}} \\
Qwen3-4B            & 164 & 88  & 417 & 604 \\
Qwen3-8B            & 257 & 390 & 369 & 554 \\
Qwen3-14B           & 298 & 76  & 163 & 252 \\
Qwen3-32B           & 157 & 154 & 146 & 246 \\
Gemma-3-4B          & 167 & 128 & 398 & 783 \\
Gemma-3-12B         & 153 & 192 & 167 & 226 \\
Gemma-3-27B         & 311 & 350 & 217 & 251 \\
\bottomrule
\end{tabular}
\caption{Mean latency and tokens per task of the task-level and subtask-level agents for each model, averaged over the None, Text, and Code formats and pooled over the three benchmarks. Tokens count the prompt and completion tokens of all model calls in a task.}
\label{tab:cost}
\end{table}

\subsection{Cost Metrics}
\label{app:eval:cost}
The latency of a task is its wall-clock time. The dependency is the MEM1 cost \citep{zhou2026mem}, which approximates the attention work a task spends on its context. For a task whose model calls each have an input length $p_c$ and an output length $o_c$, it is
\begin{equation*}
\mathrm{dependency} = \sum_{c} \frac{(2 p_c + o_c)\, o_c}{2},
\end{equation*}
reported in units of $10^{9}$ token$^2$. Both costs are read from the execution logs and never steer the agents.

% Per-model cost, compared only within one model.
\Table~\ref{tab:cost} reports the latency and tokens per task of each model. We compare them only within one model, because latency also depends on the provider and the hardware. The subtask-level agent spends more tokens per task than the task-level agent on every model, $1.2$ to $2.2$ times as many. Its latency is also higher on the four API models but not on every local model, so \Figure~\ref{fig:efficiency} also reports dependency, which depends on the tokens alone.

% Total cost of the study.
The whole study, including all reruns in this appendix, runs $123{,}777$ tasks with $33.1$B prompt tokens and $0.80$B completion tokens, which add up to about $7{,}800$ hours of per-task wall-clock time.

\begin{figure}[t]
\centering
\includegraphics[width=\linewidth]{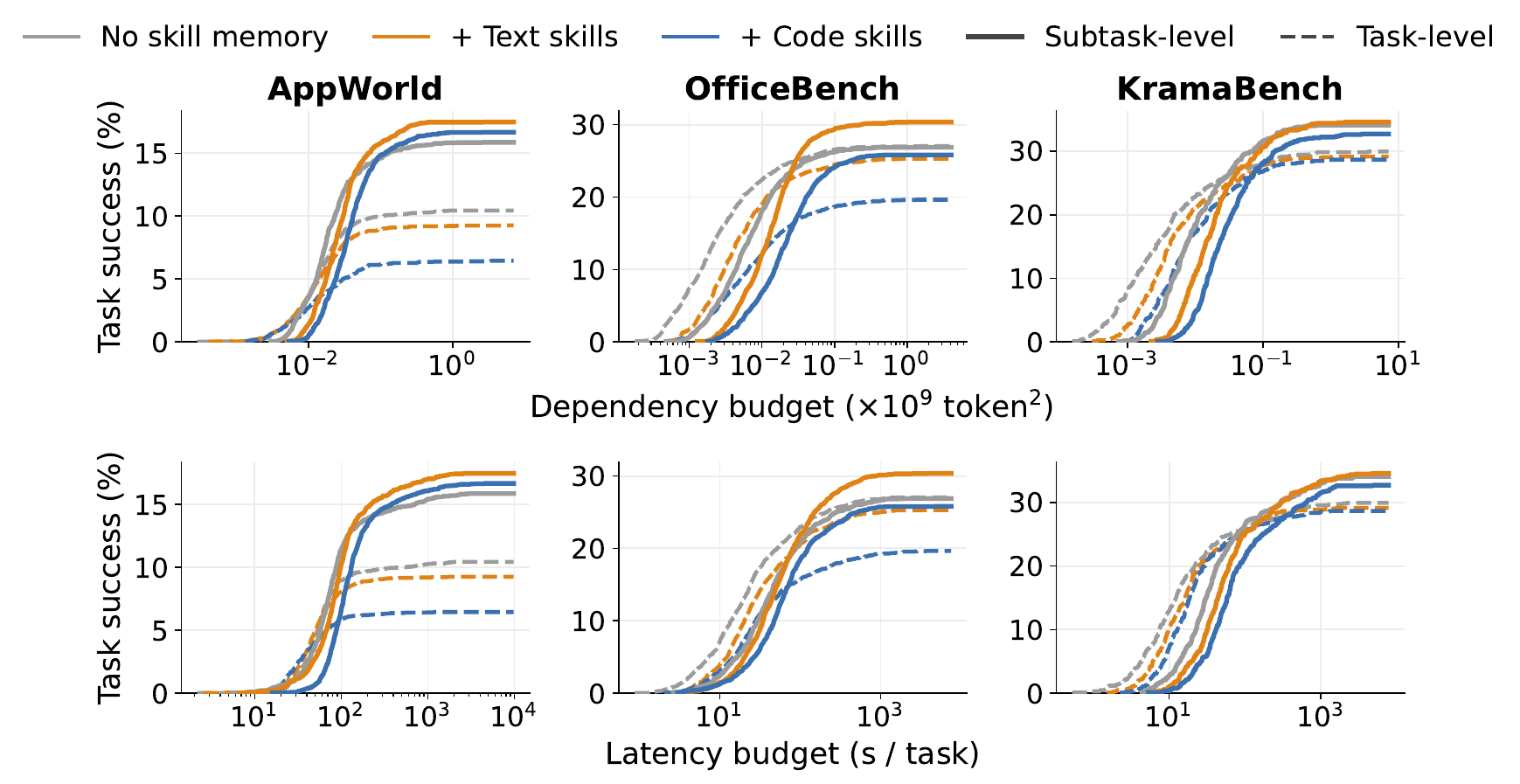}
\caption{Task success reachable within a per-task budget of dependency (left column) and latency (right column), one row per benchmark, pooled over models, with color the skill format and line style the induction level as in \Figure~\ref{fig:efficiency}. The crossover of the main text appears on every benchmark, where the task-level agent leads at small budgets and the subtask-level agent overtakes and saturates higher.}
\label{fig:budget-bench}
\end{figure}

\begin{table*}[t!]
\centering
\scriptsize \setlength{\tabcolsep}{3pt} \renewcommand{\arraystretch}{0.95}
\resizebox{\textwidth}{!}{%

\begin{tabular}{ll ccc ccc}
\toprule
 & & \multicolumn{3}{c}{Task-level} & \multicolumn{3}{c}{Subtask-level} \\
\cmidrule(lr){3-5}\cmidrule(lr){6-8}
Benchmark & Model & None & +Text & +Code & None & +Text & +Code \\
\midrule
\multirow{12}{*}{AppWorld}
 & Qwen3-235B-A22B     & 7.4\ci{5.0}{10.1} & 18.0\ci{14.4}{21.8} & 14.4\ci{11.0}{18.0} & 27.3\ci{23.0}{31.7} & 35.5\ci{30.9}{40.0} & \textbf{35.7}\ci{31.2}{40.3} \\
 & GPT-OSS-120B        & \textbf{27.3}\ci{23.3}{31.7} & 17.0\ci{13.4}{20.6} & 1.0\ci{0.2}{1.9} & 26.4\ci{22.3}{30.5} & 25.2\ci{21.1}{29.5} & 23.7\ci{19.7}{27.8} \\
 & Nemotron-Super-120B & 8.9\ci{6.2}{11.8} & 4.3\ci{2.4}{6.2} & 1.4\ci{0.5}{2.6} & 17.7\ci{14.1}{21.6} & \textbf{21.6}\ci{17.7}{25.4} & 18.5\ci{14.9}{22.3} \\
 & Qwen3-4B            & 0.0\ci{0.0}{0.0} & 0.2\ci{0.0}{0.7} & 0.0\ci{0.0}{0.0} & 0.2\ci{0.0}{0.7} & 0.7\ci{0.0}{1.7} & \textbf{1.4}\ci{0.5}{2.6} \\
 & Qwen3-8B            & 0.2\ci{0.0}{0.7} & 0.0\ci{0.0}{0.0} & 0.0\ci{0.0}{0.0} & 2.2\ci{1.0}{3.6} & \textbf{3.1}\ci{1.7}{5.0} & 1.9\ci{0.7}{3.4} \\
 & Qwen3-14B           & 0.5\ci{0.0}{1.2} & 0.7\ci{0.0}{1.7} & 0.2\ci{0.0}{0.7} & 6.7\ci{4.3}{9.4} & 7.2\ci{4.8}{9.8} & \textbf{8.4}\ci{5.8}{11.0} \\
 & Qwen3-32B           & 1.9\ci{0.7}{3.4} & 2.9\ci{1.4}{4.6} & 1.4\ci{0.5}{2.6} & 7.7\ci{5.3}{10.3} & \textbf{10.3}\ci{7.4}{13.2} & 7.2\ci{4.8}{9.8} \\
 & Gemma-3-4B          & \textbf{0.0}\ci{0.0}{0.0} & \textbf{0.0}\ci{0.0}{0.0} & \textbf{0.0}\ci{0.0}{0.0} & \textbf{0.0}\ci{0.0}{0.0} & \textbf{0.0}\ci{0.0}{0.0} & \textbf{0.0}\ci{0.0}{0.0} \\
 & Gemma-3-12B         & 0.0\ci{0.0}{0.0} & 0.2\ci{0.0}{0.7} & 0.2\ci{0.0}{0.7} & \textbf{1.7}\ci{0.5}{3.1} & 1.4\ci{0.5}{2.6} & 1.0\ci{0.2}{1.9} \\
 & Gemma-3-27B         & 0.5\ci{0.0}{1.2} & 0.5\ci{0.0}{1.2} & 0.0\ci{0.0}{0.0} & 4.6\ci{2.6}{6.7} & 4.3\ci{2.4}{6.5} & \textbf{4.8}\ci{2.9}{7.0} \\
 & Gemini-3.1-Pro      & 68.1\ci{63.5}{72.4} & 58.0\ci{53.2}{62.6} & 52.3\ci{47.5}{57.1} & 68.3\ci{63.8}{72.7} & 72.4\ci{68.1}{76.7} & \textbf{77.5}\ci{73.4}{81.5} \\
\cmidrule(lr){2-8}
 & Average                & 10.4\ci{9.6}{11.3} & 9.3\ci{8.4}{10.2} & 6.5\ci{5.8}{7.1} & 14.8\ci{13.5}{16.2} & \textbf{16.5}\ci{15.1}{18.0} & 16.4\ci{15.1}{17.7} \\
\cmidrule(lr){1-8}
\multirow{12}{*}{OfficeBench}
 & Qwen3-235B-A22B     & \textbf{43.0}\ci{37.3}{48.7} & 38.3\ci{33.0}{44.0} & 36.7\ci{31.3}{42.0} & 38.7\ci{33.3}{44.3} & 41.3\ci{35.7}{47.0} & 38.7\ci{33.3}{44.0} \\
 & GPT-OSS-120B        & 32.3\ci{27.0}{37.7} & 29.3\ci{24.3}{34.3} & 5.0\ci{2.7}{7.7} & 38.0\ci{32.7}{43.3} & \textbf{42.0}\ci{36.7}{47.7} & 40.7\ci{35.3}{46.3} \\
 & Nemotron-Super-120B & 28.3\ci{23.3}{33.7} & 31.7\ci{26.3}{37.0} & 12.7\ci{9.0}{16.7} & 27.3\ci{22.3}{32.3} & \textbf{37.7}\ci{32.3}{43.3} & 25.0\ci{20.0}{30.0} \\
 & Qwen3-4B            & 17.7\ci{13.3}{22.0} & 16.3\ci{12.3}{20.7} & 13.0\ci{9.3}{17.0} & 18.0\ci{13.7}{22.7} & \textbf{25.3}\ci{20.7}{30.3} & 20.3\ci{16.0}{25.0} \\
 & Qwen3-8B            & \textbf{25.0}\ci{20.0}{30.0} & 15.7\ci{11.7}{20.0} & 16.3\ci{12.3}{20.7} & 19.0\ci{14.7}{23.7} & \textbf{25.0}\ci{20.3}{30.0} & 24.7\ci{19.7}{29.7} \\
 & Qwen3-14B           & 28.3\ci{23.3}{33.7} & 24.0\ci{19.3}{29.0} & 27.3\ci{22.3}{32.3} & 27.0\ci{22.0}{32.0} & \textbf{32.0}\ci{27.0}{37.3} & 30.3\ci{25.3}{35.7} \\
 & Qwen3-32B           & 34.3\ci{29.0}{39.7} & 32.7\ci{27.3}{38.0} & 35.3\ci{30.0}{41.0} & 33.3\ci{28.3}{38.7} & \textbf{36.0}\ci{30.7}{41.7} & 31.7\ci{26.7}{37.0} \\
 & Gemma-3-4B          & \textbf{3.7}\ci{1.7}{6.0} & 3.0\ci{1.3}{5.0} & 2.3\ci{0.7}{4.3} & 2.7\ci{1.0}{4.7} & 1.0\ci{0.0}{2.3} & 0.7\ci{0.0}{1.7} \\
 & Gemma-3-12B         & 10.0\ci{6.7}{13.7} & 19.7\ci{15.3}{24.3} & 11.0\ci{7.7}{14.7} & 13.7\ci{10.0}{17.7} & \textbf{21.3}\ci{17.0}{26.0} & 20.0\ci{15.7}{24.7} \\
 & Gemma-3-27B         & \textbf{28.3}\ci{23.3}{33.7} & 26.7\ci{21.7}{31.7} & 14.7\ci{10.7}{18.7} & 24.0\ci{19.0}{29.0} & 19.0\ci{14.7}{23.7} & 23.7\ci{19.0}{28.7} \\
 & Gemini-3.1-Pro      & 46.3\ci{40.7}{52.0} & 41.0\ci{35.3}{46.7} & 41.7\ci{36.3}{47.3} & 47.3\ci{41.7}{53.0} & 46.0\ci{40.3}{51.7} & \textbf{48.7}\ci{43.0}{54.3} \\
\cmidrule(lr){2-8}
 & Average                & 27.0\ci{23.6}{30.4} & 25.3\ci{22.1}{28.8} & 19.6\ci{17.0}{22.4} & 26.3\ci{23.0}{29.7} & \textbf{29.7}\ci{26.2}{33.2} & 27.7\ci{24.2}{31.2} \\
\cmidrule(lr){1-8}
\multirow{12}{*}{KramaBench}
 & Qwen3-235B-A22B     & 52.8\ci{43.0}{62.5} & 49.3\ci{39.2}{59.2} & 51.5\ci{41.1}{61.2} & 52.4\ci{42.5}{62.2} & \textbf{55.3}\ci{45.4}{64.7} & 52.7\ci{42.9}{62.7} \\
 & GPT-OSS-120B        & 31.7\ci{22.8}{41.1} & 28.4\ci{19.6}{37.4} & 22.5\ci{14.2}{31.1} & 48.1\ci{38.3}{58.2} & 49.8\ci{39.6}{59.5} & \textbf{50.9}\ci{40.8}{60.6} \\
 & Nemotron-Super-120B & 53.3\ci{43.2}{63.2} & 52.1\ci{42.1}{62.2} & 43.3\ci{33.5}{53.4} & \textbf{59.8}\ci{50.0}{69.1} & 57.6\ci{47.6}{67.6} & 49.9\ci{40.1}{59.7} \\
 & Qwen3-4B            & 8.3\ci{3.3}{14.4} & 12.3\ci{6.2}{19.0} & 13.3\ci{6.9}{20.4} & 13.8\ci{7.5}{20.8} & \textbf{15.1}\ci{8.3}{22.5} & 11.5\ci{5.5}{18.1} \\
 & Qwen3-8B            & 10.7\ci{4.8}{17.2} & \textbf{15.5}\ci{8.7}{22.9} & 14.9\ci{8.3}{22.0} & 14.4\ci{7.7}{21.5} & 11.2\ci{5.4}{17.7} & 14.1\ci{7.5}{21.5} \\
 & Qwen3-14B           & 20.1\ci{12.5}{28.2} & 14.9\ci{8.5}{22.0} & 23.1\ci{14.7}{31.6} & 19.1\ci{11.5}{27.1} & \textbf{23.5}\ci{15.2}{32.5} & 20.3\ci{12.5}{28.6} \\
 & Qwen3-32B           & 30.0\ci{21.1}{39.2} & 25.8\ci{17.6}{34.8} & 30.2\ci{21.4}{39.3} & 34.3\ci{25.0}{43.9} & \textbf{39.0}\ci{29.3}{48.6} & 38.9\ci{29.2}{48.7} \\
 & Gemma-3-4B          & 1.5\ci{0.0}{4.0} & 1.1\ci{0.0}{3.3} & 1.9\ci{0.0}{4.9} & \textbf{6.0}\ci{1.6}{11.4} & 4.3\ci{1.0}{8.6} & 2.2\ci{0.0}{5.4} \\
 & Gemma-3-12B         & 16.3\ci{9.7}{23.7} & 14.0\ci{7.3}{21.2} & 11.8\ci{6.0}{18.2} & \textbf{17.9}\ci{10.7}{25.7} & 17.7\ci{10.6}{25.3} & 14.2\ci{7.9}{21.3} \\
 & Gemma-3-27B         & 19.7\ci{12.2}{27.7} & 22.0\ci{14.0}{30.6} & 18.6\ci{11.3}{26.5} & \textbf{25.3}\ci{16.7}{34.4} & 23.9\ci{15.9}{32.7} & 24.2\ci{16.0}{33.1} \\
 & Gemini-3.1-Pro      & 74.3\ci{65.6}{82.5} & 74.1\ci{65.3}{82.4} & 75.1\ci{66.4}{83.3} & \textbf{75.2}\ci{66.5}{83.2} & 73.7\ci{64.9}{82.0} & 72.4\ci{63.6}{80.8} \\
\cmidrule(lr){2-8}
 & Average                & 29.0\ci{24.1}{34.0} & 28.1\ci{23.2}{33.3} & 27.8\ci{23.0}{32.9} & 33.3\ci{28.0}{38.7} & \textbf{33.7}\ci{28.5}{39.1} & 31.9\ci{26.8}{37.1} \\
\midrule
\multicolumn{2}{l}{Average (11 models)} & 22.1\ci{20.2}{24.2} & 20.9\ci{18.9}{22.9} & 18.0\ci{16.1}{19.9} & 24.8\ci{22.7}{27.0} & \textbf{26.7}\ci{24.5}{28.8} & 25.3\ci{23.3}{27.4} \\
\bottomrule
\end{tabular}
}
\caption{Task success (\%) of all eleven models, the full version of \Table~\ref{tab:main} with Gemma-3-4B and Gemma-3-12B included. Brackets are 95\% task-bootstrap confidence intervals. Each in-block Average row is taken over the models on that benchmark, and the bottom row over all models and benchmarks.}
\label{tab:main-full}
\end{table*}

\section{Additional Analyses of Skill Induction}
\label{app:extra}

\subsection{Full Results of All Models}
\label{app:full-table}

% The two weakest models rejoin the main table here; every quoted average already includes them.
\Table~\ref{tab:main-full} extends \Table~\ref{tab:main} with the two weakest models, Gemma-3-4B and Gemma-3-12B. Gemma-3-4B stays at or near the floor on all three benchmarks, and Gemma-3-12B stays at the floor on AppWorld. Every Average row and every number quoted in the main text already includes both models. \Figure~\ref{fig:budget-bench} reports efficiency cost on all benchmarks.

% This appendix provides supporting analyses for the main results. \Figure~\ref{fig:prompt-ablation} shows that the two headline comparisons of \Section~\ref{sec:results} survive weaker induction prompts, and \Figure~\ref{fig:tradeoff} shows the specificity-abstractness tradeoff behind the skill utility score. All figures use the same common task subset as the main results.

\begin{figure}[h]
\centering
\includegraphics[width=\linewidth]{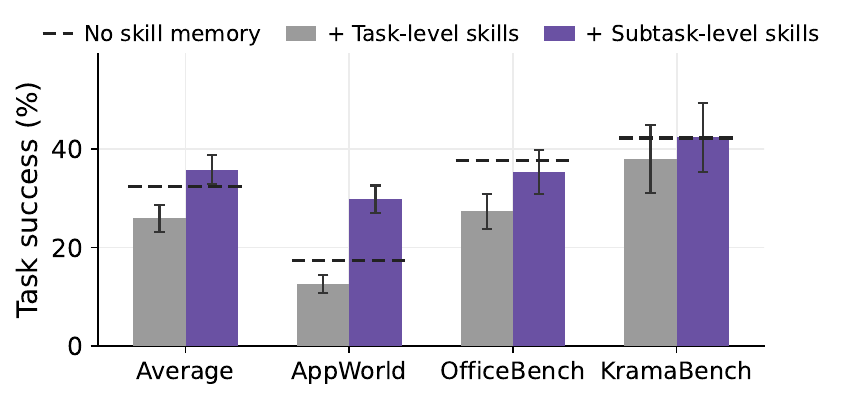}
\caption{Task success of the task-level agent retrieving its original task-level skills or the subtask-level agent's skills, averaged over the Text and Code formats and pooled over Qwen3-235B-A22B and GPT-OSS-120B, with 95\% task-bootstrap confidence intervals. Dashes mark the task-level agent without skill memory. With its original skills the task-level agent falls below the dashes on every benchmark, while with subtask-level skills it beats the original skills everywhere and returns to or above the dashes.}
\label{fig:linear-subskills}
\end{figure}

\subsection{Cases Outside the Main Trends}
\label{app:exceptions}

% When task-level skills help.
Task-level skills help mainly where the no-memory baseline is very low, as for Qwen3-235B-A22B on AppWorld at $7.4\%$ in \Table~\ref{tab:main}. Across the 33 model and benchmark cells, their effect shrinks as the baseline rises, with Spearman $\rho=-0.55$ for Text and $-0.45$ for Code ($p<0.05$).

% When subtask-level skills fail to help.
Subtask+Text falls below its no-memory baseline in only two difficulty strata of \Figure~\ref{fig:difficulty}, OfficeBench easy and KramaBench hard, by $1.9$ and $0.4$ points. Both paired 95\% intervals cover zero, so the skills neither help nor hurt there.

% When task-level code skills derail the agent.
Task-level code skills drop GPT-OSS-120B on AppWorld from $27.3\%$ to $1.0\%$ (\Table~\ref{tab:main}). In $69\%$ of these tasks the agent ends within two steps, against $0.2\%$ without memory, because it calls the retrieved whole-task function once and stops. With subtask-level code skills, only $0.2\%$ of its tasks end this early.

% When the format effect reverses.
For Gemini-3.1-Pro, Code skills beat Text skills at the subtask level on AppWorld and OfficeBench (\Table~\ref{tab:main}). Across the eleven models, model capability does not predict the format gap, as the no-memory success does not predict the gap between Code and Text skills ($\rho=0.23$, $p=0.47$). The reversal is thus specific to Gemini-3.1-Pro.

\subsection{Induction-Level Effect}
\label{app:swap}
\label{app:skill-source}

\SmallHeading{Task-level agent with subtask skills}
% Memory swap: the gain follows the skills, not the agent.
We verify that the gap between the two induction levels comes from the induced skills rather than from the agent that uses them. We replay the skill memory induced by the subtask-level agent into the task-level agent at retrieval time, so the task-level agent enters each task with the same library the subtask-level agent had at that point, and we freeze induction so the library stays the subtask-level agent's own. The task-level agent keeps its own retrieval and receives about as many skills per task from the subtask-level skill library as from its own task-level skill library, specifically $3.0$, $3.1$, and $2.0$ skills against $3.1$, $3.1$, and $1.8$ skills on AppWorld, OfficeBench, and KramaBench.

% Result: subtask skills beat the original skills everywhere and recover the harm.
\Figure~\ref{fig:linear-subskills} compares the task-level agent retrieving subtask skills against the same agent retrieving its original task-level skills, averaged over the two skill formats, on Qwen3-235B-A22B and GPT-OSS-120B. Subtask-level skills beat the original skills on every benchmark, by $9.9$ points on average and up to $17.2$ points on AppWorld. The original skills drag the task-level agent below its no-memory baseline on every benchmark, while the subtask skills return it to the level of that baseline on OfficeBench and KramaBench and lift it $12.4$ points above on AppWorld. The same agent thus turns from harmed to helped once its memory holds subtask-level skills, so the effect of the induction level travels with the skills.

\SmallHeading{Outcomes of skill source tasks}
Skill induction is not gated on task outcome in either agent. For each induced skill we therefore check whether the task it was induced from was solved in that run. Pooled over all models and benchmarks, the share of skills induced from unsolved tasks is $74.7\%$ for Task+Text, $83.0\%$ for Task+Code, $75.1\%$ for Subtask+Text, and $76.6\%$ for Subtask+Code. The two levels draw on source tasks of similar outcomes under both formats, so the opposite skill effects in \Table~\ref{tab:main} are not explained by one level distilling more failed experience than the other.

\subsection{Run-to-Run Variance}
\label{app:shuffle}

% Design: a full rerun with a reshuffled task order.
The confidence intervals in \Table~\ref{tab:main} capture task-sampling variance but not the variance across runs. We therefore rerun all six conditions on Qwen3-235B-A22B, GPT-OSS-120B, Nemotron-Super-120B, and Qwen3-32B with the task stream reshuffled.
% Result: every effect keeps its sign.
Paired per task against the same agent without memory, task-level skills change success by $-1.3$ and $-1.6$ points with Text in the original and the reshuffled run, and by $-8.6$ and $-6.9$ points with Code. Subtask-level skills change it by $+3.8$ and $+3.6$ points with Text, and by $+2.0$ and $+2.5$ points with Code. All four effects keep their sign and size, and the effects of single models and benchmarks correlate at $r=0.71$ between the two runs.

\begin{figure}[t]
\centering
\includegraphics[width=\linewidth]{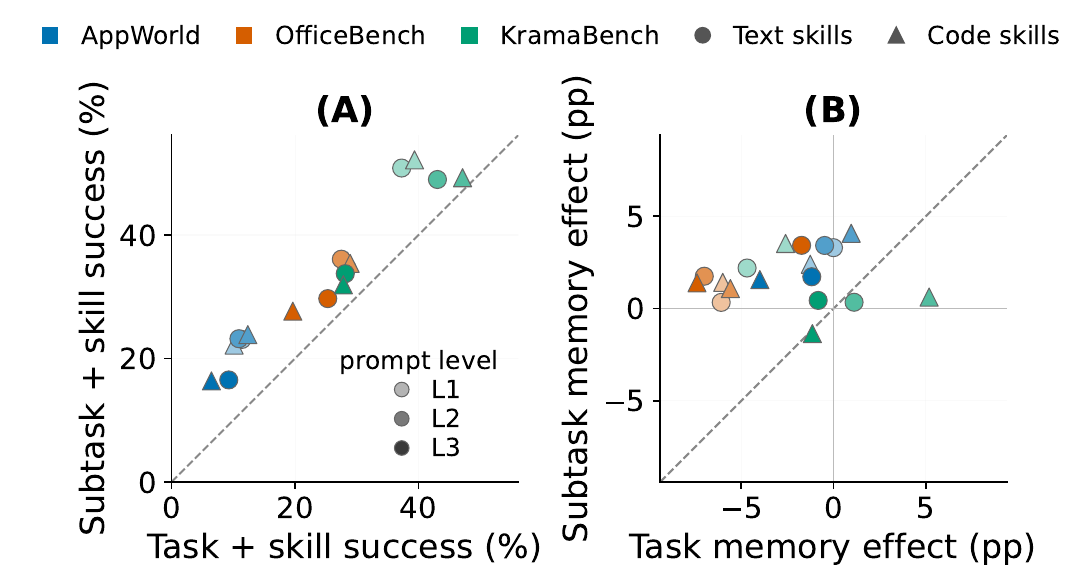}
\caption{Subtask-level induction stays ahead of task-level induction at every induction-prompt level. Each point is one benchmark, skill format, and prompt level, where L3 is the deployed prompt and L1 and L2 are the reduced versions of \Section~\ref{sec:experiments}, averaged over the models that ran it. Panel (A) compares the task success of the two skill arms, and panel (B) compares each arm's memory effect against its own no-memory baseline on the same tasks. Points above the dashed diagonal favor the subtask level, and the upper left quadrant of (B) holds the points where task-level memory hurts while subtask-level memory helps.}
\label{fig:prompt-ablation}
\end{figure}

% Second rerun: reduced induction prompts.
We also rerun both induction levels on the same four models with the two reduced induction prompts of \Section~\ref{sec:experiments}, a minimal instruction (L1) and the instruction plus one demonstration (L2). \Figure~\ref{fig:prompt-ablation} compares the two levels for each benchmark, skill format, and prompt level. The subtask-level agent with skills beats the task-level agent with skills in all $18$ cases, so the advantage of subtask-level induction does not depend on the induction prompt.

\section{Additional Analyses of Skill Utility}

\begin{figure}[t]
\centering
\includegraphics[width=\linewidth]{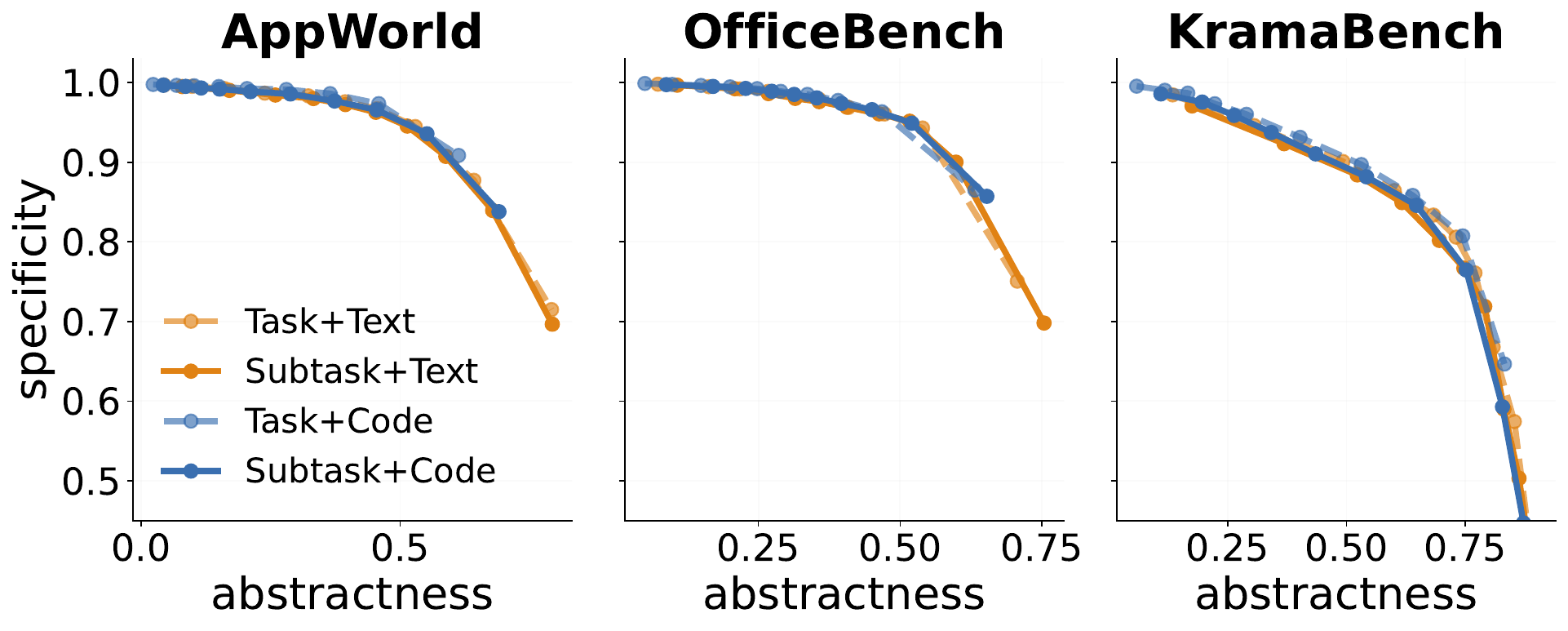}
\caption{The tradeoff between the two terms of the skill utility score. Each panel bins the skills of one benchmark into equal-size abstractness deciles and plots mean specificity against mean abstractness for the four skill conditions. All conditions fall on nearly one frontier, and subtask induction trades a small loss in specificity for a large gain in abstractness, which raises the product.}
\label{fig:tradeoff}
\end{figure}

\subsection{Specificity and Abstractness Tradeoff}
\label{app:utility-tradeoff}
\Figure~\ref{fig:tradeoff} shows the tradeoff between specificity and abstractness in the skill utility score. As specificity goes up, abstractness declines, and vice versa. We show in \Figure~\ref{fig:score-quality}b that the success of the subtask-level agent rises and then falls as specificity or abstractness alone increases, since a skill high on one dimension is low on the other. Thus, a useful skill has to balance specificity and abstractness, which the product of both rewards.

\subsection{Robustness of Skill Utility}
\label{app:utility-robust}

\SmallHeading{Skill utility with a different embedder}
% Embedder: the score does not rely on the retrieval embedder.
The score uses the same embedder as the retriever, so a high score could reflect how easily the retriever finds a skill. We therefore recompute the score with \texttt{all-mpnet-base-v2}~\cite{NEURIPS2020_c3a690be}, an embedder that never serves retrieval, on all eleven models. Over 10 equal-size bins of retrieved utility, the Spearman correlation with task success is $0.79$ for the task-level agent and $0.95$ for the subtask-level agent ($p<0.01$), against $0.98$ and $0.95$ with the retrieval embedder.

\SmallHeading{Skill utility with different aggregation}
We combine the two terms with a product because a reusable skill must meet both requirements. Any form that requires both terms behaves alike. Over 5 equal-size bins of retrieved utility, the sum, the harmonic mean, and the product all give a Spearman correlation of $1.00$ with task success for both agents, and their success rates differ by at most $0.7$ points in any bin. We use the product because the two requirements form a conjunction.

% Robustness: the score reads a property of the skills, not of the tasks they run on.
\SmallHeading{Skill utility with different difficulty}
The score is also stable across different task difficulty. Across each benchmark's native difficulty levels, the mean utility of retrieved skills stays nearly constant, moving only from $0.250$ to $0.257$ on AppWorld, from $0.294$ to $0.277$ on OfficeBench, and from $0.362$ to $0.321$ on KramaBench, pooled over all models, both skill formats, and both agents (\Table~\ref{tab:utility-difficulty}).

\subsection{Causal Effect of Skill Utility}
\label{app:utility-causal}
% Design: intervene on the library, hold everything else fixed.
We test the causal effect of skill utility by intervening on the skill library while holding everything else fixed. We split each library at its median skill utility into two complementary halves and rerun the task-level agent on the same $417$ Appworld, $300$ OfficeBench, and $92$ KramaBench tasks with only one half in memory, using the per-task replay of \Appendix~\ref{app:swap} with induction frozen, for both the task-level and the subtask-level library. Retrieval is same top-5 with 0.3 threshold, so the two arms run the same tasks and receive the same number of skills per task, and differ only in which skills they receive.

\begin{table}[t]
\centering
\small
\setlength{\tabcolsep}{4pt}
\begin{tabular}{llcc}
\toprule
Benchmark & Level & High-utility half & Low-utility half \\
\midrule
AppWorld    & Task    & \textbf{15.4}\ci{12.8}{19.1} & 10.2\ci{8.1}{12.2} \\
            & Subtask & \textbf{30.5}\ci{27.8}{33.4} & 29.6\ci{26.8}{32.6} \\
OfficeBench & Task    & \textbf{30.3}\ci{25.9}{34.8} & 26.0\ci{22.0}{30.2} \\
            & Subtask & \textbf{34.0}\ci{29.6}{38.4} & 31.5\ci{27.4}{35.7} \\
KramaBench  & Task    & \textbf{38.3}\ci{31.1}{45.5} & 36.9\ci{29.8}{43.9} \\
            & Subtask & \textbf{46.7}\ci{39.3}{54.2} & 45.8\ci{38.3}{53.4} \\
\bottomrule
\end{tabular}
\caption{Task success (\%) of the task-level agent retrieving only the high-utility or only the low-utility half of the same skill library, split at the median skill utility, on Appworld, OfficeBench, and KramaBench, averaged over the Text and Code formats and pooled over Qwen3-235B-A22B and GPT-OSS-120B, with 95\% task-bootstrap confidence intervals. Both arms run the same tasks with the same number of retrieved skills per task, and the high-utility half wins for both libraries.}
\label{tab:utility-halves}
\end{table}

% Result: the high half wins at both granularities.
\Table~\ref{tab:utility-halves} shows the result, averaged over the Text and Code formats and pooled over Qwen3-235B-A22B and GPT-OSS-120B. The high-utility half outperforms the low-utility half for both the task-level library in all of the six cases. The score, computed from skill descriptions and task instructions alone, thus identifies in advance the half of a library that produces more successes on the same tasks.

\begin{table}[t]
\centering
\small
\begin{tabular}{llcc}
\toprule
Benchmark & Difficulty & Task score & Retrieved utility \\
\midrule
\multirow{3}{*}{AppWorld}    & Level 1 & 0.215 & 0.250 \\
                             & Level 2 & 0.115 & 0.258 \\
                             & Level 3 & 0.082 & 0.257 \\
\midrule
\multirow{3}{*}{OfficeBench} & 1 app   & 0.320 & 0.294 \\
                             & 2 apps  & 0.353 & 0.286 \\
                             & 3 apps  & 0.107 & 0.277 \\
\midrule
\multirow{2}{*}{KramaBench}  & Easy    & 0.451 & 0.362 \\
                             & Hard    & 0.240 & 0.321 \\
\bottomrule
\end{tabular}
\caption{Mean task score and mean utility of the retrieved skills at each benchmark's native difficulty levels, pooled over all models, both skill formats, and both agents. The task score varies severalfold across the levels while the utility of retrieved skills stays nearly constant.}
\label{tab:utility-difficulty}
\end{table}

\begin{table*}[t]
\centering
\resizebox{\textwidth}{!}{
\begin{tabular}{llll}
\toprule
\textbf{Artifacts/Models/Packages} & \textbf{Citation} & \textbf{Link} & \textbf{License} \\
\rowcolor{\grayColor} \multicolumn{4}{c}{\textit{Benchmarks}} \\
AppWorld & \citet{trivedi-etal-2024-appworld} & \url{https://github.com/StonyBrookNLP/appworld} & Apache-2.0 License \\
OfficeBench & \citet{wang2024officebench} & \url{https://github.com/zlwang-cs/OfficeBench} & Apache-2.0 License \\
KramaBench & \citet{lai2025kramabench} & \url{https://github.com/mitdbg/Kramabench} & MIT License (Code) and CC-BY-NC-4.0 (data) \\
\rowcolor{\grayColor} \multicolumn{4}{c}{\textit{Backbone Models}} \\
Qwen3 (4B, 8B, 14B, 32B, 235B-A22B) & \citet{yang2025qwen3} & \url{https://huggingface.co/collections/Qwen/qwen3} & Apache-2.0 License \\
GPT-OSS-120B & \citet{agarwal2025gpt} & \url{https://huggingface.co/openai/gpt-oss-120b} & Apache-2.0 License \\
Nemotron-3-Super-120B & \citet{nvidia2026nemotron3superopen} & \url{https://huggingface.co/nvidia/NVIDIA-Nemotron-3-Super-120B-A12B-BF16} & NVIDIA Open Model License \\
Gemma-3 (4B, 12B, 27B) & \citet{gemmateam2025gemma3technicalreport} & \url{https://deepmind.google/models/gemma/gemma-3/} & Gemma Terms of Use \\
Gemini-3.1-Pro & N/A & \url{https://ai.google.dev/gemini-api/docs/models/gemini-3.1-pro-preview} & Missing \\
all-MiniLM-L6-v2 & \citet{reimers-gurevych-2019-sentence} & \url{https://huggingface.co/sentence-transformers/all-MiniLM-L6-v2} & Apache-2.0 License \\
\rowcolor{\grayColor} \multicolumn{4}{c}{\textit{Packages}} \\
vLLM & \citet{10.1145/3600006.3613165} & \url{https://github.com/vllm-project/vllm} & Apache-2.0 License \\
boto3 & N/A & \url{https://github.com/boto/boto3} & Apache-2.0 License \\
google-genai & N/A & \url{https://github.com/googleapis/python-genai} & Apache-2.0 License \\
Sentence-Transformers & \citet{reimers-gurevych-2019-sentence} & \url{https://github.com/huggingface/sentence-transformers} & Apache-2.0 License \\
NumPy & \citet{harris2020array} & \url{https://numpy.org} & BSD 3-Clause License \\
pandas & \citet{mckinney-proc-scipy-2010} & \url{https://pandas.pydata.org} & BSD 3-Clause License \\
SciPy & \citet{2020SciPy-NMeth} & \url{https://scipy.org} & BSD 3-Clause License \\
Matplotlib & \citet{Hunter:2007} & \url{https://matplotlib.org} & Matplotlib License (PSF-based) \\
Tesseract OCR & \citet{1288165.1288167} & \url{https://github.com/tesseract-ocr/tesseract} & Apache-2.0 License \\
LibreOffice & N/A & \url{https://www.libreoffice.org} & Mozilla Public License 2.0 \\
\bottomrule
\end{tabular}
}
\caption{Benchmarks, backbone models, and major software packages used in our study, with their citations, links, and licenses. We will release the codebase and the induced skill libraries of this project under the MIT License to support open science and reproducibility.}
\label{tab:tools}
\end{table*}

\subsection{Transfer Density and Reuse}
\label{app:density}
\label{app:reuse-format}

% Memory size: count transfer per induced skill.
\SmallHeading{Divide out the memory size}
The subtask-level agent stores more skills than the task-level agent, and a larger library can carry a transfer in more bin pairs. We therefore also count, for each induced skill, the later bins that retrieve it, which divides out the library size. Averaged over models and both formats, a subtask-level skill reaches $2.4$, $3.4$, and $3.2$ later bins on AppWorld, OfficeBench, and KramaBench, against $2.1$, $2.9$, and $1.4$ for a task-level skill, so subtask-level skills also transfer more per skill.

% \subsection{Reuse Frequency and Skill Format}
\SmallHeading{Reuse frequency and skill format}
% \label{app:reuse-format}
\Figure~\ref{fig:transfer} pools the two skill formats on purpose, because reuse frequency tracks the induction-level effect but not the format effect. Code skills are retrieved as often as Text skills, or more often, on five of the six benchmark and format combinations, yet they bring smaller gains (\Section~\ref{sec:results:fmt}). Reuse frequency measures how often a skill is retrieved, not how broadly it applies, so it separates the two induction levels but not the two formats. We therefore rank skill memories with the skill utility score, which reads the format effect off abstractness, rather than with raw reuse counts.

\section{Responsible NLP Research}
\label{app:checklist}

\subsection{Artifacts}
\label{app:checklist:artifacts}

% What we release and under which license.
To foster reproducibility and open science, we will release our complete codebase and the induced skill libraries of all conditions under the MIT License. \Table~\ref{tab:tools} details the essential artifacts of this project, covering the benchmarks, the backbone models, and the major software packages. We adhere to the intended use of every artifact, and each license permits non-commercial research applications.

% Language coverage, PII, and offensive content.
All benchmark tasks and all induced skills are in English. The released artifacts contain no personally identifiable information, because AppWorld and OfficeBench run on fully simulated user data and KramaBench draws on public scientific files. For the same reason the artifacts contain no offensive content.

\subsection{Usage of AI Assistants}
\label{app:checklist:ai}

% The three roles AI assistants play in this project.
We use Artificial Intelligence (AI) assistants in three roles, a judge inside one validation experiment, coding support, and writing support.

\SmallHeading{AI Judge}
The retrieval-quality check of \Appendix~\ref{app:selfretr} uses Gemini-3.1-Pro to judge whether sampled induced skills are relevant to their source tasks. This judge is part of the reported experiments rather than of the paper production, and its setup and results appear there.

\SmallHeading{AI Code Completions}
To streamline the development process, we leveraged \href{https://claude.com/claude-code}{Claude Code} for assistance. The tool was primarily used to generate routine code, such as inline comments, function header documentation, and boilerplate statements like \texttt{if \_\_name\_\_ == "\_\_main\_\_":}. The high-level software architecture and the core logic of all functions were manually designed and implemented by the authors.

\SmallHeading{Grammar Checking}
The initial draft of this paper was composed manually. For refinement, we employed a suite of AI-powered writing assistants, including \href{https://www.deepl.com/translator}{DeepL} for translation, and \href{https://claude.com/claude-code}{Claude Code}, \href{https://gemini.google.com/}{Gemini} for grammatical correctness.

\clearpage

\begin{figure*}[t]
\begin{promptsystem}
\promptfile{appworld_react_system.txt}
\end{promptsystem}
\caption{Execution core on AppWorld, the system prompt shared by both agents. Each subtask-level role appends its role block of \Figure~\ref{fig:prompt-subtask-aw}, and the task-level agent uses the core unchanged.}
\label{fig:prompt-core-aw}
\end{figure*}

\begin{figure*}[t]
\begin{promptsystem}
\promptfile{officebench_react_system.txt}
\end{promptsystem}
\caption{Execution core on OfficeBench, the system prompt shared by both agents. Each subtask-level role appends its role block of \Figure~\ref{fig:prompt-subtask-ob}, and the task-level agent uses the core unchanged.}
\label{fig:prompt-core-ob}
\end{figure*}

\begin{figure*}[t]
\begin{promptsystem}
\promptfile{kramabench_react_system.txt}
\end{promptsystem}
\caption{Execution core on KramaBench, the system prompt shared by both agents. Each subtask-level role appends its role block of \Figure~\ref{fig:prompt-subtask-kb}, and the task-level agent uses the core unchanged.}
\label{fig:prompt-core-kb}
\end{figure*}

\clearpage

\begin{figure*}[t]
\begin{promptsystem}[System (executor role block)]
\promptfile{appworld_executor_role.txt}
\end{promptsystem}
\begin{promptsystem}[System (planner role block)]
\promptfile{appworld_planner_role.txt}
\end{promptsystem}
\begin{promptsystem}[System (summarizer role block)]
\promptfile{appworld_summarizer_role.txt}
\end{promptsystem}
\caption{Role blocks of the subtask-level agent on AppWorld. Each block is appended to the execution core of \Figure~\ref{fig:prompt-core-aw} to form the system prompt of its role.}
\label{fig:prompt-subtask-aw}
\end{figure*}

\begin{figure*}[t]
\begin{promptsystem}[System (executor role block)]
\promptfile{officebench_executor_role.txt}
\end{promptsystem}
\begin{promptsystem}[System (planner role block)]
\promptfile{officebench_planner_role.txt}
\end{promptsystem}
\begin{promptsystem}[System (summarizer role block)]
\promptfile{officebench_summarizer_role.txt}
\end{promptsystem}
\caption{Role blocks of the subtask-level agent on OfficeBench. Each block is appended to the execution core of \Figure~\ref{fig:prompt-core-ob} to form the system prompt of its role.}
\label{fig:prompt-subtask-ob}
\end{figure*}

\begin{figure*}[t]
\begin{promptsystem}[System (executor role block)]
\promptfile{kramabench_executor_role.txt}
\end{promptsystem}
\begin{promptsystem}[System (planner role block)]
\promptfile{kramabench_planner_role.txt}
\end{promptsystem}
\begin{promptsystem}[System (summarizer role block)]
\promptfile{kramabench_summarizer_role.txt}
\end{promptsystem}
\caption{Role blocks of the subtask-level agent on KramaBench. Each block is appended to the execution core of \Figure~\ref{fig:prompt-core-kb} to form the system prompt of its role.}
\label{fig:prompt-subtask-kb}
\end{figure*}

\clearpage

\begin{figure*}[t]
\begin{promptuser}[User (executor)]
\promptfile{executor_user.txt}
\end{promptuser}
\begin{promptuser}[User (planner)]
\promptfile{planner_user.txt}
\end{promptuser}
\begin{promptuser}[User (summarizer)]
\promptfile{summarizer_user.txt}
\end{promptuser}
\caption{User messages of the three subtask-level roles. The progress summary starts as a fixed no-progress line, and the live action menu appears only on OfficeBench.}
\label{fig:prompt-subtask-user}
\end{figure*}

\begin{figure*}[t]
\begin{promptuser}[User (retrieved text skill framing)]
\promptfile{text_memory_frame.txt}
\end{promptuser}
\begin{promptuser}[User (retrieved code skill framing)]
\promptfile{code_memory_frame.txt}
\end{promptuser}
\caption{Framing wrapped around the text and code skills retrieved for the linear and subtask-level agent.}
\label{fig:prompt-subtask-frame}
\end{figure*}

\clearpage

\begin{figure*}[t]
\begin{promptuser}
\promptfile{appworld_text_induction.txt}
\end{promptuser}
\caption{Text-skill induction prompt on AppWorld, appended as a user message to the span being distilled. The same prompt serves both induction levels.}
\label{fig:prompt-text-aw}
\end{figure*}

\begin{figure*}[t]
\begin{promptuser}
\promptfile{officebench_text_induction.txt}
\end{promptuser}
\caption{Text-skill induction prompt on OfficeBench, appended as a user message to the span being distilled. The same prompt serves both induction levels.}
\label{fig:prompt-text-ob}
\end{figure*}

\begin{figure*}[t]
\begin{promptuser}
\promptfile{kramabench_text_induction.txt}
\end{promptuser}
\caption{Text-skill induction prompt on KramaBench, appended as a user message to the span being distilled. The same prompt serves both induction levels.}
\label{fig:prompt-text-kb}
\end{figure*}

\clearpage

\begin{figure*}[t]
\begin{promptuser}
\promptfile{appworld_code_induction.txt}
\end{promptuser}
\caption{Code-skill induction prompt on AppWorld, appended as a user message to the span being distilled. The same prompt serves both induction levels.}
\label{fig:prompt-code-aw}
\end{figure*}

\begin{figure*}[t]
\begin{promptuser}
\promptfile{officebench_code_induction.txt}
\end{promptuser}
\caption{Code-skill induction prompt on OfficeBench, appended as a user message to the span being distilled. The same prompt serves both induction levels.}
\label{fig:prompt-code-ob}
\end{figure*}

\begin{figure*}[t]
\begin{promptuser}
\promptfile{kramabench_code_induction.txt}
\end{promptuser}
\caption{Code-skill induction prompt on KramaBench, appended as a user message to the span being distilled. The same prompt serves both induction levels.}
\label{fig:prompt-code-kb}
\end{figure*}

\clearpage

\begin{figure*}[t]
\begin{skillbox}[Task-Level Text Skill]
\promptfile{skill_example_task_text.txt}
\end{skillbox}
\caption{The task-level text skill induced from the example task. One workflow note spans the whole trajectory from login to order placement.}
\label{fig:skill-example-task-text}
\end{figure*}

\begin{figure*}[t]
\begin{skillbox}[Task-Level Code Skill]
\promptfile{skill_example_task_code.txt}
\end{skillbox}
\caption{The task-level code skill induced from the same task. One function reproduces the whole trajectory and keeps the source task's product keyword as its default argument.}
\label{fig:skill-example-task-code}
\end{figure*}

\begin{figure*}[t]
\begin{skillbox}[Subtask-Level Text Skills]
\promptfile{skill_example_subtask_text.txt}
\end{skillbox}
\caption{The subtask-level text skills induced from the same task, one note per completed subtask.}
\label{fig:skill-example-subtask-text}
\end{figure*}

\begin{figure*}[t]
\begin{skillbox}[Subtask-Level Code Skills]
\promptfile{skill_example_subtask_code.txt}
\end{skillbox}
\caption{The subtask-level code skills induced from the same task, one function per completed subtask.}
\label{fig:skill-example-subtask-code}
\end{figure*}

\clearpage

\end{document}